\documentclass[journal]{IEEEtran}

\usepackage{graphicx}
\usepackage{amsfonts}
\usepackage{url}
\usepackage[table]{xcolor}
\usepackage{makecell}
\usepackage{subfigure}

\newcommand{\tao}[1]{\textcolor{black}{#1}}
\newcommand{\you}[1]{\textcolor{black}{#1}}

\newcommand{\round}[1]{\ensuremath{\lfloor#1\rceil}}

\begin{document}

\title{3D ToF LiDAR in Mobile Robotics: A Review}

\author{Tao Yang$^1$, 
  You Li$^2$,~\IEEEmembership{Member,~IEEE,}
  Cheng Zhao$^3$,
  Dexin Yao$^1$,
  Guanyin Chen$^1$,
  Li Sun$^4$,\\
  Tomas Krajnik$^5$,
  and Zhi Yan$^6$,~\IEEEmembership{Senior Member,~IEEE}
  \thanks{This work was supported by the National Natural Science Foundation of China under Grant 62103332, the Fundamental Research Funds for the Central Universities, NPU, under Grant 31020200QD045, Czech Science Foundation grant 20-27034J, and the Bourgogne-Franche-Comt\'e regional research project LOST-CoRoNa.\\ $^1$Unmanned System Research Institute, Northwestern Polytechnical University, China\\ $^2$Research division of RENAULT S.A.S, France\\ $^3$Department of Engineering Science, University of Oxford, UK\\ $^4$Research division of NIO Inc., China\\ $^5$Faculty of Electrical Engineering, Czech Technical University, Czechia\\ $^6$CIAD UMR7533, Univ. Bourgogne Franche-Comt\'e, UTBM, F-90010 Belfort, France}}

\markboth{}{Yang \MakeLowercase{\textit{et al.}}: 3D ToF LiDAR in Mobile Robotics: A Review}

\maketitle

\begin{abstract}
  In the past ten years, the use of 3D Time-of-Flight (ToF) LiDARs in mobile robotics has grown rapidly.
  Based on our accumulation of relevant research, this article systematically reviews and analyzes the use 3D ToF LiDARs in research and industrial applications.
  The former includes object detection, robot localization, long-term autonomy, LiDAR data processing under adverse weather conditions, and sensor fusion.
  The latter encompasses service robots, assisted and autonomous driving, and recent applications performed in response to public health crises.
  We hope that our efforts can effectively provide readers with relevant references and promote the deployment of existing mature technologies in real-world systems.
\end{abstract}


\section{Introduction}
\label{sec:introduction}

\IEEEPARstart{L}{iDAR} stands for Light Detection And Ranging, and as its name suggests, it is a technology (or equipment) that uses pulsed light, typically emitted by a laser, to measure distances.
Historically, there have been several waves of research on how mobile robots could perceive the outside world: sonars, planar laser rangefinders, passive and active visual sensors, and today's millimeter wave radars and 3D LiDARs.
Among them, the two-dimensional LiDAR, thanks to its ability to provide accurate geometric representation of the environment, has significantly influenced efficiency of mobile robots, by enabling robust and accurate metric localisation and mapping (i.e. SLAM)~\cite{probabilistic_robotics}.

\begin{figure}[t]
  \centering
  \includegraphics[width=\columnwidth]{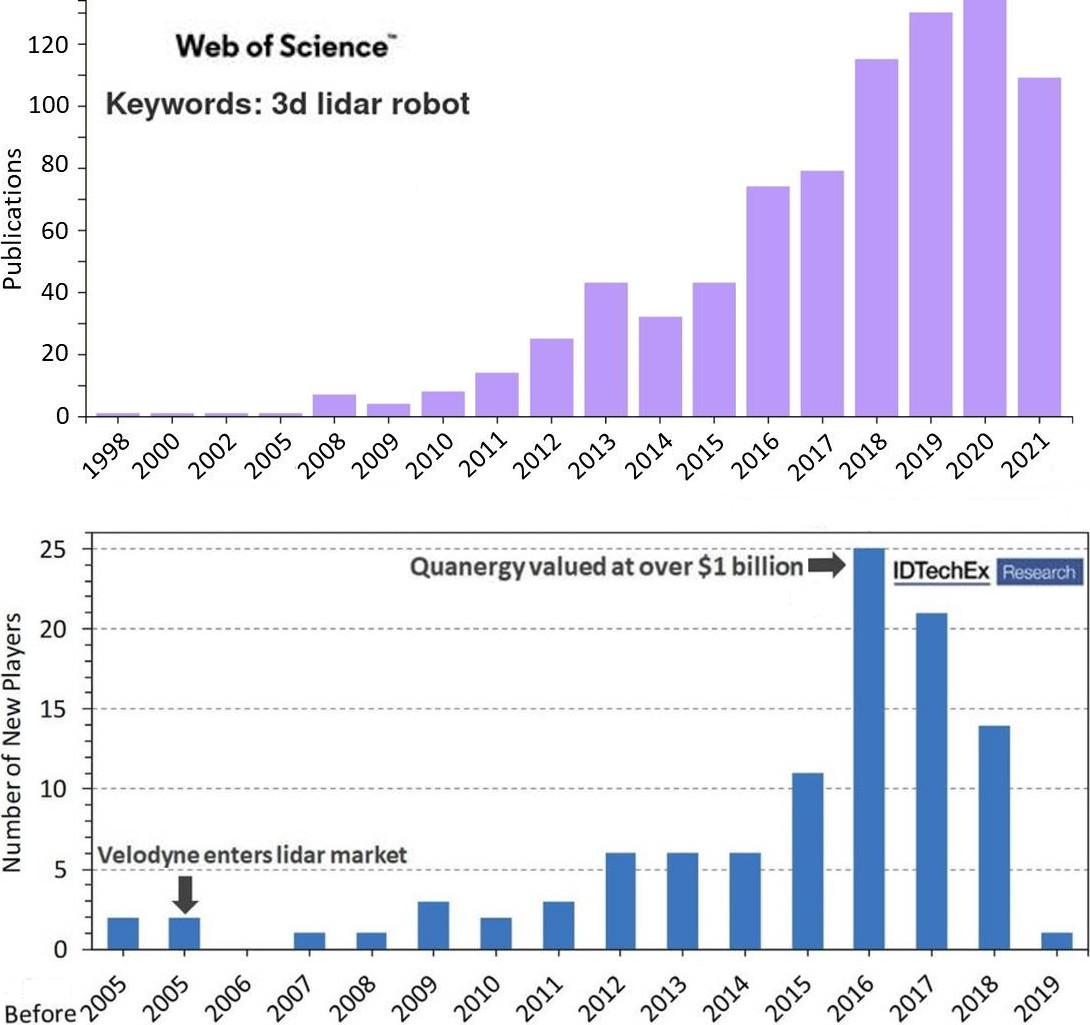}
  \caption{Upper: the results of the number of relevant papers per year searched on the Web of Science with ''3D lidar robot'' as the keyword. Lower: new LiDAR manufacturers per year through 2019~\cite{idtechex}.}
  \label{fig:papers-players}
\end{figure}

\begin{figure}[t]
  \centering
  \includegraphics[width=\columnwidth]{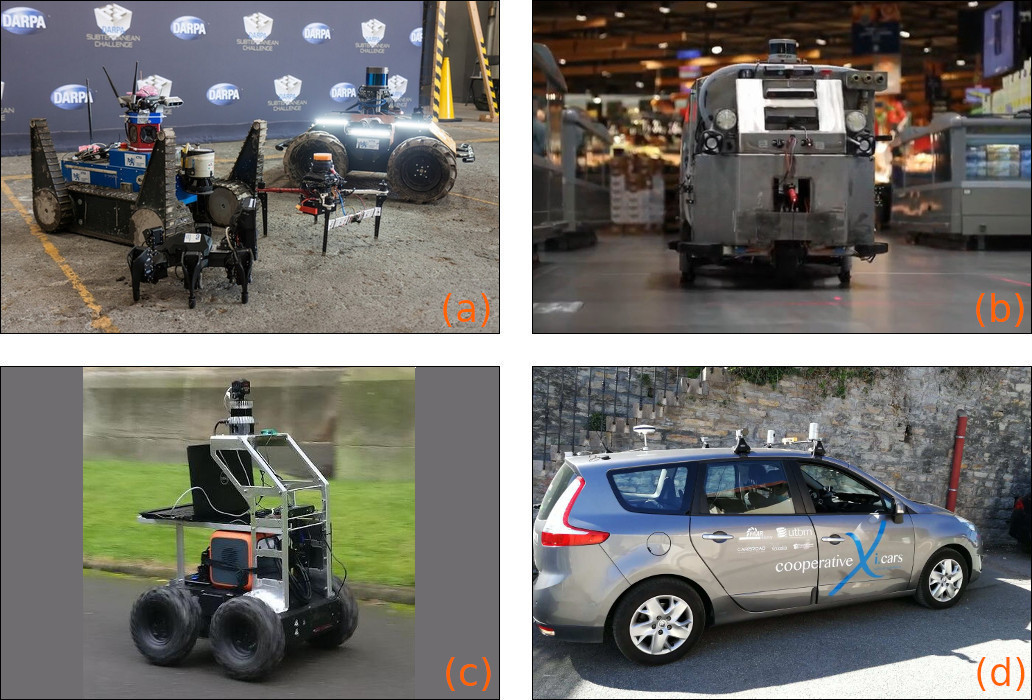}
  \caption{Various LiDARs used in our projects.
    (a) A RoboSense RS-LiDAR-32 3D LiDAR used by the CTU-CRAS team in the DARPA Sub-T Challenge~\cite{DARPA1}.
    (b) A professional cleaning robot equipped with a Velodyne VLP-16 3D LiDAR~\cite{zhimon20jist}.
    (c) An Ouster OS1-64 3D LiDAR installed on the top of a field robot~\cite{ls21iros}.
    (d) An autonomous car equipped with two Velodyne HDL-32E 3D LiDARs~\cite{yz20iros}.}
  \label{fig:LiDAR-robot-car}
\end{figure}

Today, standing on the shoulders of giants, researchers have the opportunity to explore and study more challenging issues in more complex environments.
In the past ten years, 3D LiDAR has received more and more attention, which can be seen from the number of relevant papers published, the amount of capital investment and the number of industry players involved (see Fig.~\ref{fig:papers-players}).
Compared with classic 2D LiDARs, modern 3D LiDARs not only add a extra spatial dimension by increasing the number of scanning layers, but also achieve wider fields of view and range.
For example, the latest off-the-shelf product, e.g. Velodyne Alpha Prime, can achieve 360-degree scanning and a measurement range of up to 300 meters.
As a consequence, 3D LiDAR has been used as an important part of perception systems in mobile robotics (see Fig.~\ref{fig:LiDAR-robot-car}).

This article aims to provide readers with information on the timeliness of 3D time-of-flight (ToF) LiDAR applications in mobile robotics, to classify and analyze research in related fields, and ultimately to provide guidance for deployment of 3D LiDARs in robotic applications.
In order to provide a deeper insight in the referenced methods, this article is primarily based on hand-on experience obtained when using 3D LiDARs in the projects we participated in.
Therefore, the research axis listed in the following is neither exclusive nor exhaustive.
The contributions of this review include:
\begin{itemize}
\item An extensive literature survey of 3D LiDAR in the field of mobile robots, focusing primarily on topics related to perception while providing less detailed information to other relevant research (e.g. target tracking, SLAM, etc.).
\item A new publicly available benchmark suite for point cloud segmentation\footnote{\url{https://github.com/cavayangtao/lidar_clustering_bench}}, which is based on open source code and open datasets.
\end{itemize}

The remainder of our review is organized as follows.
Section~\ref{sec:preliminaries} introduces background knowledge in order to make following this article easier.
Section~\ref{sec:axes} lists the research that we have focused on in recent years, including object detection, robot localization, long-term robot autonomy, LiDAR data processing under adverse weather conditions, as well as sensor fusion.
Section~\ref{sec:applications} introduces application areas, including service robots, autonomous driving, and current anti-epidemic deployments.
Section~\ref{sec:perspectives} provides a vision for future research and development. 
Section~\ref{sec:conclusions} summarizes the full text.

\section{Preliminaries}
\label{sec:preliminaries}

\subsection{Ranging Principle}

\begin{figure}[t]
  \centering
  \includegraphics[width=\columnwidth]{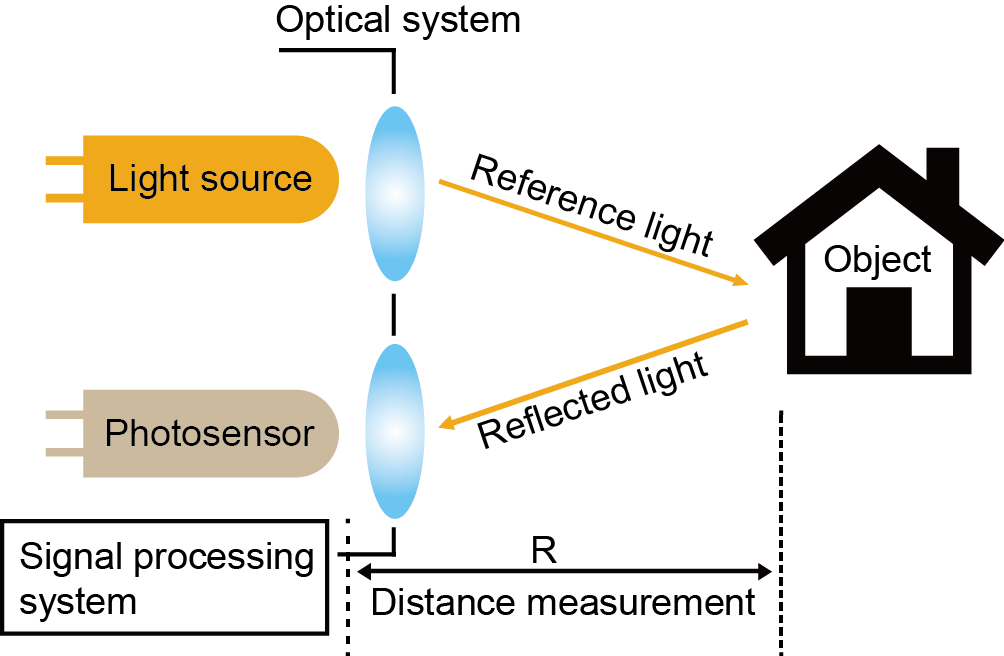}
  \caption{Schematic diagram of the imaging principle of ToF LiDAR.}
  \label{fig:LiDAR-principles}
\end{figure}

As mentioned earlier, LiDAR is a surveying method that measures distance to a target by illuminating the target with pulsed laser light and measuring the reflected pulses with a sensor.
Fig.~\ref{fig:LiDAR-principles} shows the working principle of ToF LiDAR.
Specifically, the laser transmitter first emits pulsed laser light in a given direction, and when the laser beam encounters an obstacle, it will reflect or diffuse depending on the material on the surface of the object.
After the laser detector receives the echo signal, it can determine the distance between the sensor and the object by measuring the time it takes for the laser beam to reach the object from the sensor and then back again.
This measurement mechanism can be formulated as:
\begin{equation}
  r=\frac{c \cdot \Delta t}{2 n}
  \label{range_eq}
\end{equation}
where $c$ is the speed of light, which is a universal physical constant; $n$ is the refractive index of the propagation medium, which is 1 for air; $\Delta T$ is the time difference between transmitting and receiving laser pulses.

\begin{figure}[t]
  \centering
  \includegraphics[width=\columnwidth]{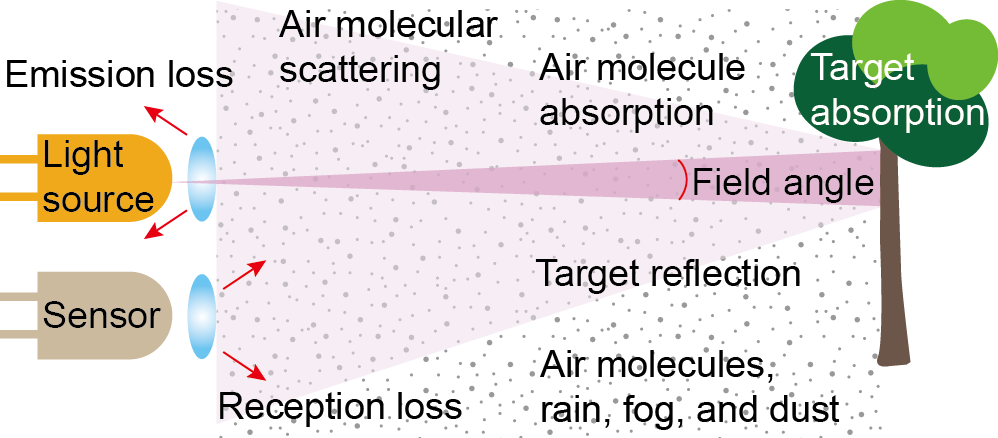}
  \caption{Difficult laser return journey.}
  \label{fig:laser_return}
\end{figure}

The reception of a laser beam is not as straightforward as its emission (see Fig.~\ref{fig:laser_return}).
The received laser power function is described by the backscattering coefficient, which can be formulated as~\cite{Rasshofer2011}:
\begin{equation}\label{eq:rashof}
  P_{R}=C \frac{\beta}{R^{2}} \exp \left(-2 \int_{r=0}^{R} \alpha(r) d r\right)
  \label{power_eq}
\end{equation}
where $P_R$ is the power of a received laser return at distance $R$;
$C$ is the constant related to the light speed, laser transmit power, laser detector's optical aperture area, and overall system efficiency;
$\beta$ is the reflection efficiency of the object surface;
$\alpha$ is the extinction coefficient of the LiDAR signal.
The advanced signal processing system is required to detect the true return signal in low SNR (Signal-to-Noise Ratio).
A typical final stage of the processing pipeline is adaptive thresholding, which takes into account the current background noise~\cite{Ogawa2016}.

\subsection{Physical Structure}

\begin{figure}[t]
  \centering
  \includegraphics[width=\columnwidth]{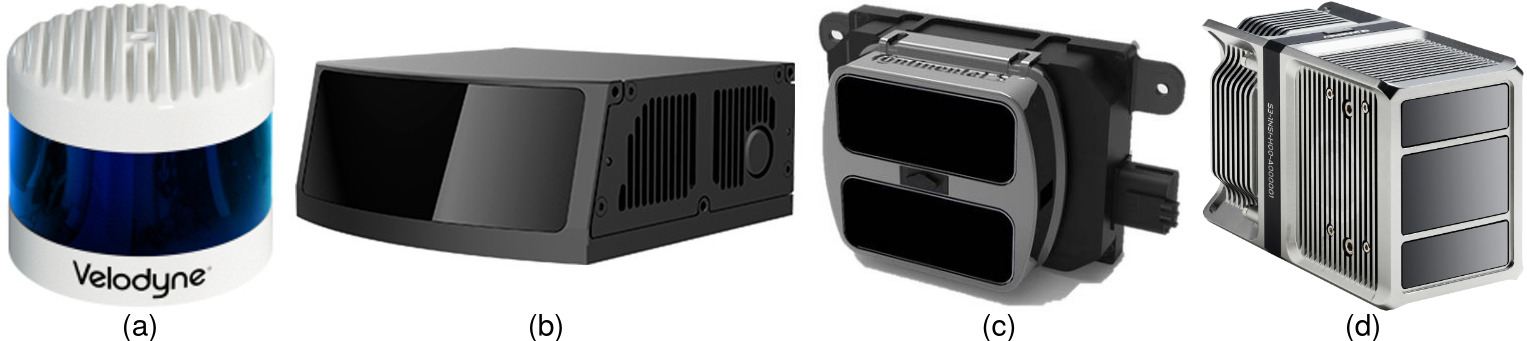}
  \caption{Example products of different types of 3D ToF LiDAR on the market.
    (a) Velodyne Alpha Prime spanning LiDAR.
    (b) Robosense M1 MEMS LiDAR.
    (c) Continental HFL110 3D Flash LiDAR.
    (d) Quanergy S3-2 OPA LiDAR.}
  \label{fig:LiDAR-types}
\end{figure}

The 3D ToF LiDAR on the market today can be divided into four categories (see Fig.~\ref{fig:LiDAR-types}) according to the different physical structure of its scanning system, including mechanical rotation, Micro-Electro-Mechanical Systems (MEMS), flash, and Optical Phased Array (OPA), of which the latter three are also called solid-state LiDAR.
The mechanical rotation system utilizes a motor to rotate the mirror of the laser beam to obtain a large horizontal Field of View (FoV).
By stacking transmitters in the vertical direction or an embedded oscillating lens, lasers in the vertical FoV are generated.
MEMS LiDAR uses the electromagnetic force to rotate a micro mirror integrated in the chip to generate dense points at high frequencies in a specific field.
It is even possible to control the circuit to generate dense points concentrated in certain specific areas.
Flash LiDAR instantly illuminates the scene through the lens.
A photodiode array (similar to CMOS) is used to receive the reflected signal to generate a set of 3D points recorded by each pixel.
OPA LiDAR realizes scanning in different directions by changing the phase difference of each unit of the laser source array.
This type is still in the experimental stage so far.
Table~\ref{tab:scanning} summarizes the advantages and disadvantages of different scanning systems for mass-produced LiDARs. 
For a detailed physical comparison of different LiDARs, please refer to~\cite{you20spm}.

\begin{table}[t]
  \caption{Pros and cons of different categories of 3D ToF LiDAR}
  \label{tab:scanning}
  \begin{center}
    \begin{tabular}{|l|l|l|}
      \hline
      \textbf{Structure} & \textbf{Pros} & \textbf{Cons}\\
      \hline\hline
      Spinning & High technology maturity & Limited life (by wear)\\
      & Up to 360$^{\circ}$ of horizontal FoV & Large size and high cost\\
      \hline
      MEMS & Near solid-state structure & Limited detection range\\
      & Small size and low cost & Limited FoV\\
      \hline
      Flash & Solid-state structure & Low detection range\\
      & Very small size and low cost & Low FoV\\
      \hline
    \end{tabular}
  \end{center}
\end{table}

For the popular mechanical LiDAR, achieving a higher number of scanning layers needs to stack laser transceiver modules, which means the increasing of both the cost and the difficulty of manufacture.
Thus, although the mechanical LiDARs are widely used due to their high technological maturity and 360 degree of horizontal FoV, industry and academia are increasingly investing in MEMS LiDAR, especially in the application and mass production in the field of autonomous driving.
However, the small reflection mirror and light receiving aperture limit the detection range of MEMS LiDAR, and transceiver modules and scanning modules still need to be further improved.
Flash LiDAR detects the whole area with a single diffused laser, which makes it difficult to achieve a large detection range due to the limitation of laser emission power on human eye safety.
Therefore, it is usually used in mid-distance or indoor scenes.

\subsection{Physical Characteristics}

An important specification of LiDAR is the ability to provide fast data collection and high-precision distance measurement.
In addition, mechanical rotation-based devices, which are widely used in mobile robotics, can also provide large-scale scans, with up to 360 degrees of horizontal FoV and a measurement distance of tens to hundreds of meters.
Furthermore, LiDARs are robust to lighting variations, which makes them more suitable for long-term robot autonomy compared to passive visual sensors.
However, the data provided by LiDAR come in a form of a sparse set of points, making objects difficult to identify due to the lack of easy-to-interpret features such as color and texture.
This situation usually becomes more serious as the measurement distance increases, because the sensory data becomes sparser with distance. 
Another limitation of LiDAR is its sensitivity to certain adverse weather conditions.
Water droplets in the air have a dual effect on the sensor.
First, small droplets in the atmosphere absorb or scatter the near-infrared laser, resulting in an increase in the extinction coefficient $\alpha$ (c.f. Eq.~\ref{eq:rashof}).
Second, wet surfaces of obstacles lead to weaker reflectivity $\beta$~\cite{Lekner1988}.
These factors cause the received laser power to be low, making it impossible to perform the signal processing steps required to detect distant objects.
In addition, raindrops and snowflakes near the laser transmitter cause noise and result in false detections~\cite{sac2022george}.
Table~\ref{tab:sensors_pros_and_cons} gives an overview of the performance of commonly used sensors for machine (exteroceptive) perception.

\begin{table}[t]
  \caption{Overview of the performance of commonly used sensors for machine (external) perception}
  \label{tab:sensors_pros_and_cons}
  \begin{center}
    \begin{tabular}{|l|l|l|l|l|}
      \hline
      \textbf{Sensor} & \textbf{LiDAR} & \textbf{Radar}& \textbf{Ultrasonic}& \textbf{Camera}\\
      \hline\hline
      Detection range & $\star\star\star$ & $\star\star\star$ & $\star$ & $\star\star$\\
      \hline
      Ranging accuracy & $\star\star\star$ & $\star\star$ & $\star\star$ & $\star$\\
      \hline
      Resolution & $\star\star$ & $\star$ & $\star$ & $\star\star\star$\\
      \hline
      Horizontal FoV & $\star\star\star$ & $\star\star$ & $\star$ & $\star\star$\\
      \hline
      Vertical FoV & $\star\star$ & $\star$ & $\star$ & $\star\star$\\
      \hline
      Color information & $\star\star$ & $\star$ & $\star$ & $\star\star\star$\\
      \hline
      Lighting robustness & $\star\star\star$ & $\star\star\star$ & $\star\star\star$ & $\star$\\
      \hline
      Weather robustness & $\star\star$ & $\star\star\star$ & $\star\star\star$ & $\star$\\
      \hline
    \end{tabular}
  \end{center}
\end{table}

\subsection{Data Representation and Processing}

LiDAR's knowledge to its surroundings can be represented by a set of points in three-dimensional coordinate space:
\begin{equation}
  P = \{p_i~|~p_i = (x_i,y_i,z_i) \in \mathbb{R}^3, i = 1, \ldots, I\}
\end{equation}
while each point corresponds to a processed laser beam reflection.
Conventionally, this set of points is called a \emph{point cloud}.
According to different characteristics of the hardware, additional features might be included, such as intensity and the laser ring number.
A point cloud processing library widely used in the field of mobile robotics and autonomous driving is the PCL (Point Cloud Library)~\cite{pcl}.
This library is open-source, written in C\texttt{++}, based on traditional computer vision algorithms (in contrast to deep neural networks) including feature estimation, surface reconstruction, 3D registration, model fitting, and segmentation.
PCD (Point Cloud Data)\footnote{\url{http://pointclouds.org/documentation/tutorials/pcd_file_format.php}} file format is the point cloud data storage format officially supported by PCL.

Another way to represent and process LiDAR data is inseparable from the well-known Robot Operating System (ROS)~\cite{ros}.
ROS has become the \emph{de facto} standard platform for development of software in robotics, and today increasing numbers of researchers and industries develop autonomous driving software based on it.
The LiDAR data is usually represented by \emph{PointCloud2} ROS message\footnote{\url{http://docs.ros.org/melodic/api/sensor_msgs/html/msg/PointCloud2.html}}, and can be easily saved and shared via \emph{rosbag}~\cite{yz17iros,yz19auro,yz20iros}.
Data processing can be completely handed over to PCL due to the close family relationship between ROS and PCL.
In addition, other representations such as binary file (e.g. .bin)~\cite{KITTI,oxford-dataset,KAIST,cadcd} and xml file~\cite{nuScenes} are also common.

\section{Research axes}
\label{sec:axes}

Table~\ref{tab:taxonomic} provides the full taxonomic axes from the literature, that will be presented in detail in the following subsections.
\begin{table*}[t]
  \caption{Summary of the taxonomic axes}
  \label{tab:taxonomic}
  \begin{center}
    \begin{tabular}{|l|l|l|l|l|l|l|l|l|}
      \hline
      \textbf{Paper} & \textbf{Axes} & \textbf{Method} & \textbf{Intensity} & \textbf{Ring} & \textbf{3D ToF LiDAR} & \textbf{Usage$^1$} & \textbf{Dataset} & \textbf{OS$^2$}\\
      \hline\hline
      Yan \emph{et al.}~\cite{zhimon20jist} & detection & clustering+SVM$^3$ & \checkmark & $\times$ & Velodyne VLP-16 & combined & FLOBOT & \checkmark\\
      \hline
      Yan \emph{et al.}~\cite{yz17iros} & detection & clustering+SVM & \checkmark & $\times$ & Velodyne VLP-16 & alone & L-CAS & \checkmark\\
      \hline
      Yang \emph{et al.}~\cite{yang21itsc} & detection & online RF$^4$ & \checkmark & $\times$ & Velodyne HDL-64E & combined & KITTI & \checkmark\\
      \hline
      Kidono \emph{et al.}~\cite{kidono11iv} & detection & clustering+SVM & \checkmark & $\times$ & Velodyne HDL-64E & alone & private & $\times$\\
      \hline
      Dewan \emph{et al.}~\cite{dewan16icra} & detection & motion curve & $\times$ & $\times$ & Velodyne HDL-64E & alone & KITTI & $\times$\\
      \hline
      Zhou and Tuzel~\cite{VoxelNet} & detection & DL$^5$ & \checkmark & $\times$ & Velodyne HDL-64E & alone & KITTI & $\times$\\
      \hline
      Ali \emph{et al.}~\cite{YOLO3D} & detection & DL & $\times$ & $\times$ & Velodyne HDL-64E & alone & KITTI & $\times$\\
      \hline
      N-Serment \emph{et al.}~\cite{navarro-serment09fsr} & detection & tracking+SVM & $\times$ & $\times$ & unknown & alone & private & $\times$\\
      \hline
      H\"aselich \emph{et al.}~\cite{haselich14iros} & detection & clustering+SVM & $\times$ & $\times$ & Velodyne HDL-64E & alone & Freiburg+private & $\times$\\
      \hline
      Li \emph{et al.}~\cite{li16its} & detection & clustering+SVM & $\times$ & $\times$ & Velodyne HDL-64E & alone & private & $\times$\\
      \hline
      Wang and Posner~\cite{wang15rss} & detection & SW$^6$+SVM & \checkmark & $\times$ & Velodyne HDL-64E & alone & KITTI & $\times$\\
      \hline
      Spinello \emph{et al.}~\cite{spinello11icra} & detection & clustering+AB$^7$ & $\times$ & $\times$ & Velodyne HDL-64E & alone & Freiburg & $\times$\\
      \hline
      Deuge \emph{et al.}~\cite{deuge13acra} & detection & clustering+SVM & \checkmark & $\times$ & Velodyne HDL-64E & alone & Sydney Urban & $\times$\\
      \hline
      Teichman and Thrun~\cite{teichman12ijrr} & detection & clustering+EM$^8$ & $\times$ & $\times$ & Velodyne HDL-64E & alone & Stanford & $\times$\\
      \hline
      Dequaire \emph{et al.}~\cite{dequaire17ijrr} & detection & DL & $\times$ & $\times$ & Velodyne HDL-64E & alone & Oxford & $\times$\\
      \hline
      Sualeh and Kim~\cite{sualeh19sensor} & detection & clustering+TM$^9$ & $\times$ & $\times$ & Velodyne HDL-64E & alone & KITTI & $\times$\\
      \hline
      Qi \emph{et al.}~\cite{qi21cvpr} & detection & DL & \checkmark & $\times$ & unkown & alone & Waymo & $\times$\\
      \hline
      Yan \emph{et al.}~\cite{yz18iros} & detection & clustering+SVM & \checkmark & $\times$ & Velodyne VLP-16 & combined & L-CAS MS & \checkmark\\
      \hline
      Sun \emph{et al.}~\cite{ls20icra} & localization & DL & $\times$ & $\times$ & Velodyne HDL-32E & alone & NCLT & $\times$\\
      \hline
      Wang \emph{et al.}~\cite{wang2020pointloc} & localization & DL & $\times$ & $\times$ & Velodyne HDL-32E & alone & Oxford & $\times$\\
      \hline
      Dub\'e \emph{et al.}~\cite{dube2017segmatch} & localization & RF+NN$^{10}$ & $\times$ & $\times$ & Velodyne HDL-64E & alone & KITTI & \checkmark\\
      \hline
      Dub\'e \emph{et al.}~\cite{segmap2018} & localization & DL+NN & $\times$ & $\times$ & Velodyne HDL-64E & alone & KITTI+private & \checkmark\\
      \hline
      Tinchev \emph{et al.}~\cite{tinchev2019learning} & localization & DL & $\times$ & $\times$ & Velodyne 64, 32, 16 & alone & KITTI+private & $\times$\\
      \hline
      Kong \emph{et al.}~\cite{kong2020semantic} & localization & DL+NN & $\times$ & $\times$ & Velodyne HDL-64E & alone & KITTI & \checkmark\\
      \hline
      He \emph{et al.}~\cite{he2016m2dp} & localization & heuristic & $\times$ & \checkmark & Velodyne HDL-64E & alone & KITTI+others & \checkmark\\
      \hline
      Kim and Kim~\cite{kim2018scan} & localization & heuristic & $\times$ & \checkmark & Velodyne 64,32,16 & alone & KITTI+others & \checkmark\\
      \hline
      Cop \emph{et al.}~\cite{cop2018delight} & localization & heuristic & \checkmark & $\times$ & Velodyne VLP-16 & alone & private & $\times$\\
      \hline
      Kim \emph{et al.}~\cite{kim20191} & localization & DL & $\times$ & \checkmark & Velodyne HDL-32E & alone & NCLT, Oxford & \checkmark \\
      \hline
      Chen \emph{et al.}~\cite{chen2020overlapnet} & localization & DL & \checkmark & $\times$ & Velodyne HDL-64E & alone & KITTI, Ford & \checkmark\\
      \hline
      Uy and Lee~\cite{angelina2018pointnetvlad} & localization & DL & $\times$ & $\times$ & Velodyne HDL-64E & alone & Oxford+private & \checkmark\\
      \hline
      Liu \emph{et al.}~\cite{liu2019lpd} & localization & DL & $\times$ & $\times$ & Velodyne HDL-64E & alone & Oxford+private & \checkmark\\
      \hline
      Kucner \emph{et al.}~\cite{kucnerconditional} & long-term & grid-based & $\times$ & $\times$ & Velodyne HDL-64E & alone & private & $\times$\\
      \hline
      Pomerleau \emph{et al.}~\cite{pomerleau2014long} & long-term & heuristic & $\times$ & $\times$ & Velodyne HDL-32E & alone & private & $\times$\\
      \hline
      Sun \emph{et al.}~\cite{ls18icra} & long-term & DL & $\times$ & $\times$ & Velodyne VLP-16 & combined & L-CAS & $\times$\\
      \hline
      Sun \emph{et al.}~\cite{ls18ral} & long-term & DL & $\times$ & $\times$ & Velodyne HDL-32E & alone & private & $\times$\\
      \hline
      Vintr \emph{et al.}~\cite{vintr19ecmr} & long-term & benchmarking & \checkmark & $\times$ & Velodyne VLP-16 & alone & L-CAS & $\times$\\
      \hline
      Vintr \emph{et al.}~\cite{vintr19icra} & long-term & hypertime & \checkmark & $\times$ & Velodyne VLP-16 & alone & L-CAS & $\times$\\
      \hline
      Vintr \emph{et al.}~\cite{vintr20iros} & long-term & benchmarking & \checkmark & $\times$ & Velodyne HDL-32E & alone & private & $\times$\\
      \hline
      Broughton \emph{et al.}~\cite{broughton2020learning} & long-term & DL+SVM & $\times$ & $\times$ & Velodyne VLP-16 & combined & private & $\times$\\
      \hline
      Rasshofer \emph{et al.}~\cite{Rasshofer2011} & weather & physical model & \checkmark & $\times$ & unknown & alone & private & $\times$\\
      \hline
      Roy \emph{et al.}~\cite{roy2020physical} & weather & physical model & \checkmark & $\times$ & unknown & alone & private & $\times$\\
      \hline
      Yang \emph{et al.}~\cite{yang20iros} & weather & GPR$^{11}$+DL & \checkmark & $\times$ & Velodyne VLP-32C & alone & private & \checkmark\\
      \hline
      Hahner \emph{et al.}~\cite{hahner2021fog} & weather & physical model & \checkmark & $\times$ & Velodyne 64, 32 & alone & DENSE/STF & \checkmark\\
      \hline
      Charron \emph{et al.}~\cite{8575761} & weather & distance-based & $\times$ & $\times$ & Velodyne VLP-32C & alone & CADC & \checkmark\\
      \hline
      Heinzler \emph{et al.}~\cite{heinzler2020cnn} & weather & DL & \checkmark & $\times$ & Velodyne VLP-32C & alone & DENSE/STF & \checkmark\\
      \hline
      JI-IL \emph{et al.}~\cite{park2020fast} & weather & intensity-based & \checkmark & $\times$ & Ouster OS-1 64 & alone & private & $\times$\\
      \hline
      Kurup \emph{et al.}~\cite{kurup2021dsor} & weather & distance-based & $\times$ & $\times$ &  unkown & alone & WADS & \checkmark\\
      \hline
    \end{tabular}
  \end{center}
      {$^1$Can be used alone or need to be fused with other modal sensors. $^2$Open Source. $^3$Support Vector Machine. $^4$Random Forest. $^5$Deep Learning.\\ $^6$Sliding Window. $^7$AdaBoost. $^8$Expectation Maximization. $^9$Template Matching. $^{10}$Nearest Neighbor. $^{11}$Gaussian Process Regression.}
\end{table*}

\subsection{Object Detection}

\begin{figure}[t]
  \centering
  \includegraphics[width=\columnwidth]{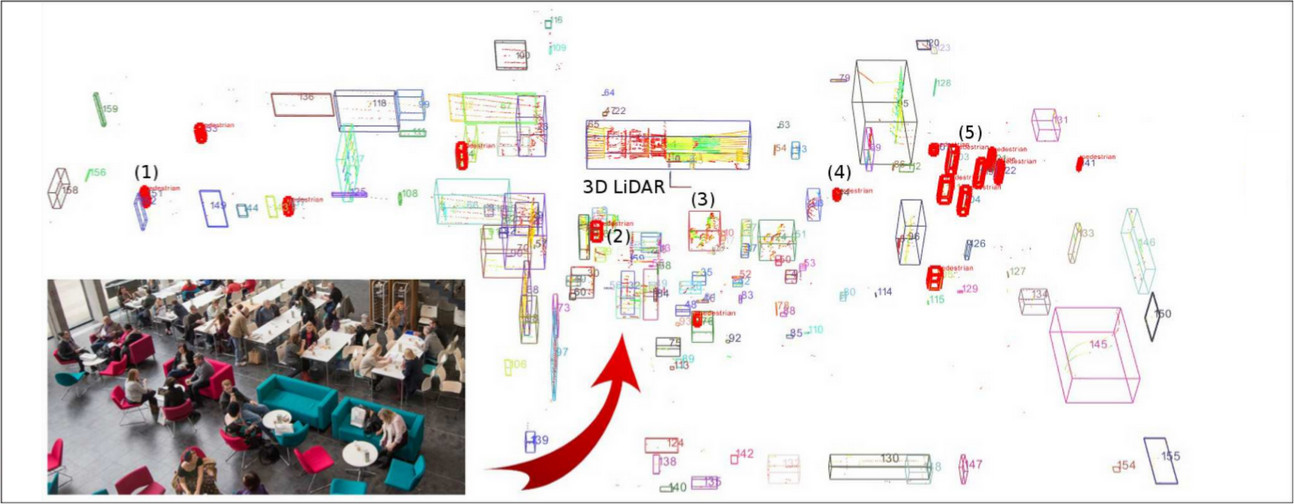}
  \includegraphics[width=\columnwidth]{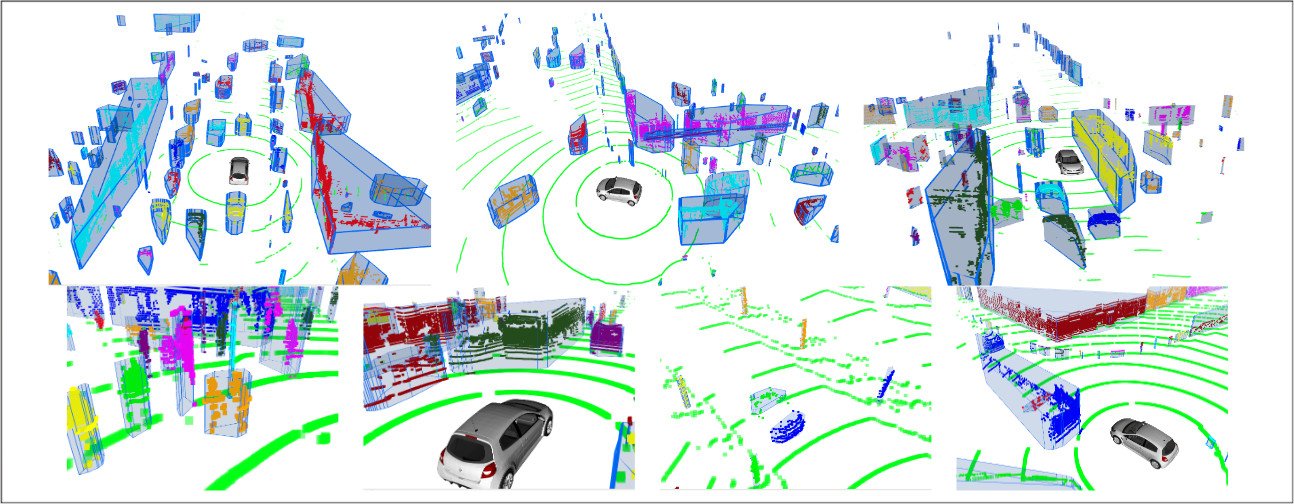}
  \caption{Examples of indoor (upper)~\cite{yz19auro} and outdoor (lower)~\cite{insclustering} object detection based on 3D LiDAR. Both are based on the segmentation-classification pipeline.}
  \label{fig:object_detection}
\end{figure}

Existing work on 3D-LiDAR-based object detection (see Fig.~\ref{fig:object_detection}) can be roughly divided into two categories, namely segmentation-classification pipeline and end-to-end pipeline.
The former first segments the point cloud into clusters~\cite{yz17iros,yz19auro,zermas17icra,bogoslavskyi16iros,autoware,insclustering} then classifies each cluster based on a given model.
This model can be based on machine learning~\cite{yz17iros,yz19auro,yz18iros,kidono11iv} or object motion~\cite{dewan16icra,shackleton10avss}.
On the other hand, the end-to-end pipeline is a modern approach and closely linked to deep learning methods, which allows to extract pedestrians and other objects directly from the point cloud~\cite{VoxelNet,YOLO3D}.
Below we conduct comparative analysis on some representative works.

\subsubsection{Segmentation-classification}

The run-based segmentation~\cite{zermas17icra} includes two steps which first extracts the ground surface in an iterative fashion using deterministically assigned seed points, and then clusters the remaining non-ground points using a two-run connected component labeling technique from binary images.
Depth clustering~\cite{bogoslavskyi16iros} is a fast method with low computational demands, which first transforms 3D LiDAR scans into 2D range images, then performs the segmentation on the latter.
The Euclidean method~\cite{RusuDissertation}, instead, clusters points by directly calculating the $L3$ distance between any two of them directly in the 3D space.
Adaptive clustering~\cite{yz17iros} is a lightweight and accurate point cloud segmentation method.
It is based on the Euclidean method~\cite{RusuDissertation} while improves the clustering performance by adding the adaptive function of the distance threshold between points.
Autoware~\cite{autoware} proposed another improved version based on~\cite{RusuDissertation}, which implicitly projects points onto a 2D (x-y) plane, and segments according to Euclidean distance.
InsClustering~\cite{insclustering} \you{further squeezes the processing time by directly process the raw data packets from a spinning LiDAR, instead of waiting the spinning LiDAR complete an entire scanning.
  The input packets are instantly segmented as ground or non-ground, then the non-ground points are clustered into clusters.
  Direct processing of the packet streams can significantly reduce the computational time required to perform the segmentation.
  It is reported that the processing time of InsClustering for a Velodyne UltraPuck LiDAR is less than 1ms, compared to 5ms by Depth clustering.}

In order to provide insight on the weaknesses and strengths of each solution, several performance comparisons of the above mentioned clustering methods were conducted.
Such comparison is essential to decide which one to adopt according to the properties of the intended final application.
However, benchmarking of InsClustering~\cite{insclustering} is not included, as it needs to be customized according to the raw data of different sensors.
All experiments reported in this article were carried out with Ubuntu 18.04 LTS (64-bit) and ROS Melodic, with an Intel i7-7700HQ processor (only one core is used), 16 GB memory, and without GPU processing.

We ran the corresponding open source code on three open datasets collected outdoors using three different LiDARs, including a Velodyne VLP-16 for the L-CAS dataset~\cite{yz19auro}, a Velodyne HDL-32E for the EU Long-term dataset~\cite{yz20iros}, and a Velodyne HDL-64E for the KITTI dataset~\cite{KITTI}.
The first one was collected in a parking lot with a stationary robot, which contains two fully labeled pedestrians with neither occlusion nor truncation on the samples.
The other two datasets were collected in urban road environments with automobile platforms, of which the EU Long-term dataset provides car labels in a roundabout, while the KITTI one includes annotations for pedestrians, cyclists, and various vehicle types.

Accurate performance evaluation is inseparable from high-quality ground-truth annotations, thus:

\begin{itemize}
\item For the L-CAS dataset, we improved the precision of annotations.
\item For the EU Long-term dataset, we extracted 200 consecutive frames from the roundabout data on April 12, 2019, and labeled all the vehicles in them.
\item For the KITTI dataset, we arbitrarily selected 200 frames from 3D object detection data and re-annotated them. This was motivated by the fact that the original bounding boxes are obtained based on RGB image projection, which contain the estimation of the full size of the vehicle and therefore cannot objectively reflect the performance of the clustering method.
\end{itemize}
Furthermore, for the latter two datasets we applied the ray ground filter~\cite{himmelsbach2010fast} to remove the ground instead of simply thresholding the z-axis, improving the integrity of the annotation bounding boxes.
All annotations were performed using the L-CAS 3D Point Cloud Annotation Tool~\cite{yz19auro}.

On the other hand, we also roughly estimated the ring number (corresponding to the scan layer of LiDAR) missed in the point cloud data of KITTI by calculating the vertical angle of a laser to meet the operating requirements of run clustering~\cite{zermas17icra} and depth clustering~\cite{bogoslavskyi16iros}:
\begin{equation}
  ring = \round{n \times \frac{\arcsin({z / \sqrt{x^2+y^2+z^2}}) + FoV_{down}} {FoV}}
  \label{eq:ring}
\end{equation}
where $n$ is the number of the LiDAR layers; $x$, $y$, and $z$ are the coordinates of a laser point; $FoV$ is the vertical field of view of the LiDAR, and $FoV_{down}$ is the vertical field of view below $0$ degree.
The final ring needs to be rounded.

First, in order to draw conclusions on the segmentation precision of the evaluated algorithms, we calculated the 3D Intersection over Union (IoU) between the clustering and the ground-truth boxes, and the experimental results are illustrated in Table~\ref{tab:clustering_bench}.
The parameter settings when the code is running are as described in the corresponding paper, while the best parameters obtained from our experiments were used in the absence of them.
It can be seen that the adaptive method performs best on the L-CAS dataset due to its distance-based segmentation of point clouds and computation of points directly in 3D space, while the depth method mainly suffers from its edge cases, i.e., objects that are too close in depth, especially background objects are larger than foreground ones.
The run method and the depth method outperform on the EU long-term and KITTI datasets, respectively, which is expected as they are more robust to uneven and sloped roads.

Additional experiments were conducted in order to further explore the clustering precision.
It is known that the ground removal based on the thresholding is mainly driven by the assumption of flat ground and the real-time requirement for the deployment of robotic systems.
However, this method requires a priori on parameter settings and suffers from uneven road surfaces.
We are therefore interested in whether effectively removing the ground can improve the performance of related methods.
As shown in the last three rows of Table~\ref{tab:clustering_bench}, by using the ray ground filter~\cite{himmelsbach2010fast}, the performance of clustering is generally improved on both EU Long-term and KITTI datasets, while it shows a negative impact on L-CAS dataset.
This is mainly because, on the one hand, the Euclidean and Autoware methods can get complete vehicle clusters in most cases, thanks to their large $\theta$ value, while the adaptive method causes the vehicle to be over-segmented due to its use of adaptive $\theta$.
On the other hand, the parking lot in the L-CAS dataset has a flat ground, while after applying the filter, some human feet are also removed, making the clusters smaller than the ground-truth in most cases.

\begin{table*}
  \caption{Parameter settings and segmentation precision of different methods on L-CAS Dataset (Velodyne VLP-16)~\cite{yz19auro}, EU Long-term Dataset (Velodyne HDL-32E)~\cite{yz20iros}, and KITTI Dataset (Velodyne HDL-64E)~\cite{KITTI}}
  \label{tab:clustering_bench}
  \begin{center}
    \begin{tabular}{|l|lll|ccc|}
      \hline
      & & Parameters & & & Precision (best in bold) &\\
      Approach & Ground removal & Min/Max points & Clustering $\theta$ & L-CAS & EU Long-term & KITTI\\

      \hline\hline
      Run clustering~\cite{zermas17icra} & $Params_{GPF}$ & 2/2.2 million & $Params_{SLR}$ & 37.63\% & \textbf{35.97\%} & 29.25\%\\
      \hline
      Depth clustering~\cite{bogoslavskyi16iros} & 7$^{\circ}$ & 5/2.2 million & 10$^{\circ}$ & 14.61\% & 28.72\% & \textbf{42.69\%}\\
      \hline
      Euclidean clustering~\cite{pcl} & -0.8m/-1.25m/-1.5m & 5/2.2 million & 0.75 m & 39.26\% & 14.78\% & 30.63\%\\
      \hline
      Adaptive clustering~\cite{yz19auro} & -0.8m/-1.25m/-1.5m & 5/2.2 million & adaptive & \textbf{62.38\%} & 32.99\% & 33.24\%\\
      \hline
      Autoware clustering~\cite{autoware} & -0.8m/-1.25m/-1.5m & 5/2.2 million & 0.75 m & 50.68\% & 34.00\% & 32.15\%\\
      \hline\hline
      Euclidean clustering~\cite{pcl} & Ray ground filter & 5/2.2 million & 0.75 m & 22.19\% & \textbf{71.12\%} & 67.16\%\\
      \hline
      Adaptive clustering~\cite{yz19auro} & Ray ground filter & 5/2.2 million & adaptive & 31.81\% & 37.13\% & 20.28\%\\
      \hline
      Autoware clustering~\cite{autoware} & Ray ground filter & 5/2.2 million & 0.75 m & 27.20\% & 62.13\% & \textbf{70.86\%}\\
      \hline
    \end{tabular}
  \end{center}
  \hspace*{1cm}{\footnotesize $Params_{GPF} = \{N_{segs} = 3, N_{iter} = 3, N_{LPR} = 20, Th_{seeds} = 0.4m, Th_{dist} = 0.2m \}$}\\
  \hspace*{1cm}{\footnotesize $Params_{SLR} = \{Th_{run} = 0.5m, Th_{merge} = 1m \}$}
\end{table*}

\begin{figure}
  \centering
  \includegraphics[width=0.83\columnwidth]{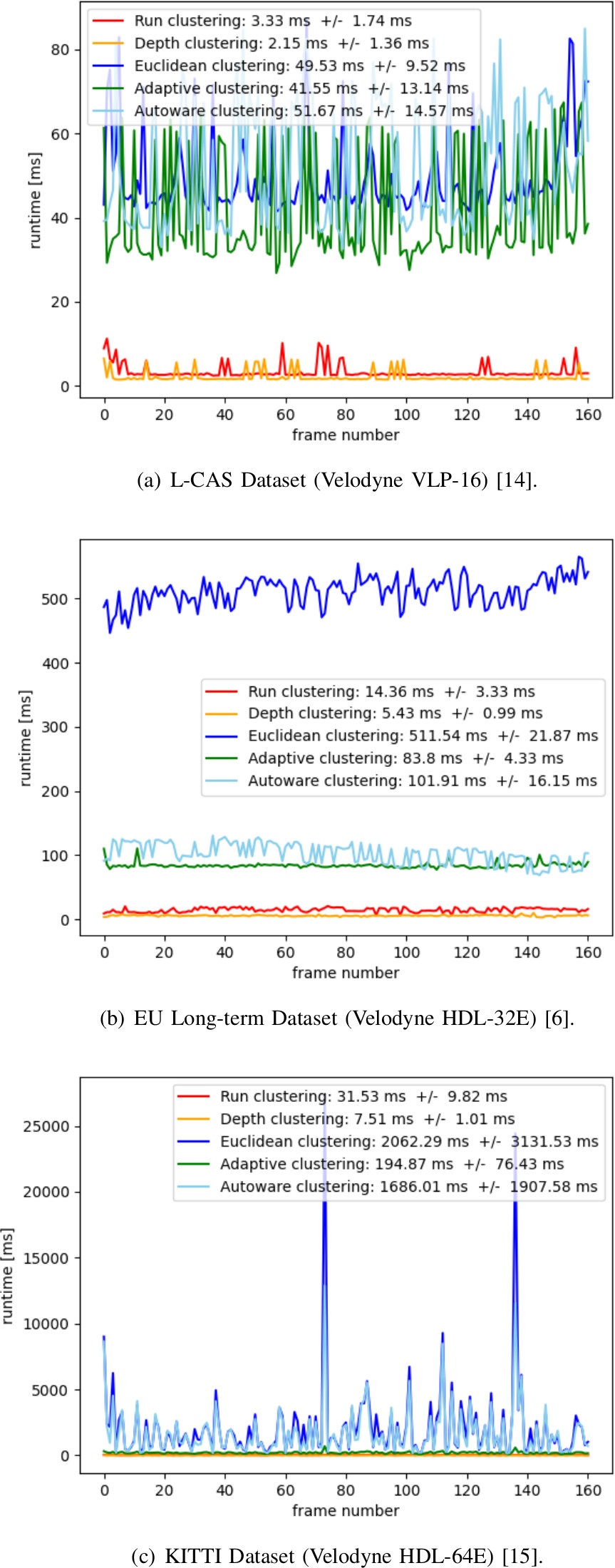}
  \caption{Run-time performance of the evaluated clustering methods.}
  \label{fig:runtime}
\end{figure}

Second, the CPU run-time of each clustering method is examined, as shown in Fig.~\ref{fig:runtime}.
It can be seen that for all methods, the processing time of the point cloud scales linearly with the number of points it contains, i.e. the shortest for 16-layer while the longest for 64-layer LiDAR.
Depth clustering has absolute performance advantages due to its dimensionality reduction processing of point cloud and industrial-grade code implementation.
Run clustering also appears to be competitive due to its clever use of priors such as ring information to avoid traversing the point cloud.
Instead, the remaining three methods are implemented based on the $k$-d tree provided by PCL with an average complexity of $\mathcal{O}(kn\log{}n)$, which is relatively time-consuming.
However, among the three, adaptive is the fastest, thanks to its distance-based division of point clouds, which actually reduces the search space of the $k$-d tree.
Autoware outperforms Euclidean with larger point clouds, because the former first projects the 3D point cloud to the 2D plane and then implements clustering, while the latter implements tree search directly in the 3D space.

Regarding object classification based on point clouds, Navarro \emph{et al.}~\cite{navarro-serment09fsr} introduced seven features for human classification and trained an SVM classifier based on these features.
Kidono \emph{et al.}~\cite{kidono11iv} proposed two additional features considering the 3D human shape and the clothing material (i.e. using the reflected laser beam intensities), showing significant classification improvements.
H{\'a}selich \emph{et al.}~\cite{haselich14iros} implemented eight of the above mentioned features for human detection in unstructured environments, discarding the intensity feature due to a lack of hardware support.
Li \emph{et al.}~\cite{li16its} implemented instead a re-sampling algorithm in order to improve the quality of the geometric features proposed by the former authors.

In terms of other research streams, Wang and Posner~\cite{wang15rss} applied a sliding window approach to 3D point data for object detection, including cars, pedestrians and cyclists.
They divided the space enclosed by a 3D bounding box into sparse feature grids, then trained an SVM classifier based on six features related to the occupancy of the cells, the distribution of points within them, and the reflectance of these points.
Spinello \emph{et al.}~\cite{spinello11icra} combined a top-down classifier, based on volumetric features, and a bottom-up detector to reduce false positives for tracking distant persons in 3D LiDAR scans.
An alternative approach by Deuge \emph{et al.}~\cite{deuge13acra} introduced an unsupervised feature learning approach for outdoor object classification by projecting 3D LiDAR scans into 2D depth images.
Recently, Qi \emph{et al.}~\cite{qi21cvpr} introduced an off-board (e.g. cloud) pipeline to get rid of the computational power limitation of on-board (i.e. robot or vehicle) methods and improve the accuracy of 3D object detection.
A key design of their detector is that different viewpoints of an object contain complementary information about its geometry within a point cloud sequence.

The above-mentioned methods are with classifier offline trained, typically under human supervision, then apply it to sensory data during robot operations.
The inconvenience with offline methods is that the pretrained classifier is not always effective when the robot is deployed for a long time (e.g. over weeks, months or years) or moves to a different environment (e.g. out-of-distribution), in which fine-tuning or retraining are typically required as a consequence.
To help relieve the user from these tedious tasks, some authors proposed methods requiring less or no annotation.
For example, Shackleton \emph{et al.}~\cite{shackleton10avss} employed a surface matching technique for human detection and an Extended Kalman Filter (EKF) to estimate the position of a human target and assist the detection in the next LiDAR scan.
Teichman \emph{et al.}~\cite{teichman12ijrr} presented a semi-supervised learning method for multi-object classification, which needs only a small set of hand-labeled seed object tracks for training the classifiers.
Dewan \emph{et al.}~\cite{dewan16icra} proposed a classifier-free approach to detect and track dynamic objects, which again relies on motion cues and is therefore not suitable for slow and static objects such as pedestrians.
Our previous work in this stream includes~\cite{yz17iros,yz18iros,yang21itsc}, in which \cite{yz17iros} presented an online learning framework for human classification in 3D LiDAR scans taking advantage of robust multi-target tracking to avoid the need for data annotation by a human expert, \cite{yz18iros} introduced a framework allowing a robot to learn a new 3D LiDAR-based human classifier from other sensors over time taking advantage of a multisensor tracking system, and \cite{yang21itsc} presented a multi-modal-based online learning system for 3D LiDAR-based object classification in urban environments, including cars, cyclists and pedestrians.

\subsubsection{End-to-end}

Emerging methods like end-to-end (typically based on deep learning) provide alternative frameworks for object detection and tracking applications.
For example, Dequaire \emph{et al.}~\cite{dequaire17ijrr} employed a Recurrent Neural Network (RNN) to capture the environment state evolution with 3D LiDAR (the 3D point clouds were reduced to a 2D scan), training a model in an entirely unsupervised manner, and then using it to track cars, buses, pedestrians and cyclists from an autonomous car.
Zhou and Tuzel~\cite{VoxelNet} developed another end-to-end network that divides the 3D point cloud into a number of voxels.
After random sampling and normalization of the points, several Voxel Feature Encoding (VFE) layers are used for each non-empty voxel to extract voxel-wise features.
This network is trained to learn an effective discriminative representation of objects with various geometries.
Simon \emph{et al.}~\cite{Complex-yolo-2018} proposed a specific Euler-Region-Proposal Network (E-RPN) to estimate the pose of an object in 3D LiDAR scans by adding an imaginary and a real fraction to the regression network.
This ends up in a closed complex space and avoids the singularities that occur with single-angle estimates, allowing robust object detection.
Yang \emph{et al.}~\cite{Pixor-2018} introduced PIXOR, a proposal-free single-stage detector that outputs oriented 3D object estimates decoded from pixel-level neural network predictions.
PIXOR takes bird's eye view representation as input for efficiency in computation.
Rencently, Lang \emph{et al.}~\cite{Pointpillars-2019} proposed another fast deep learning-based method named PointPillars, which uses PointNet~\cite{qi2017pointnet} to learn a representation of point clouds but organized in vertical columns (pillars).

\subsection{Robot Localization}

\begin{figure}[t]
  \centering
  \includegraphics[width=\columnwidth]{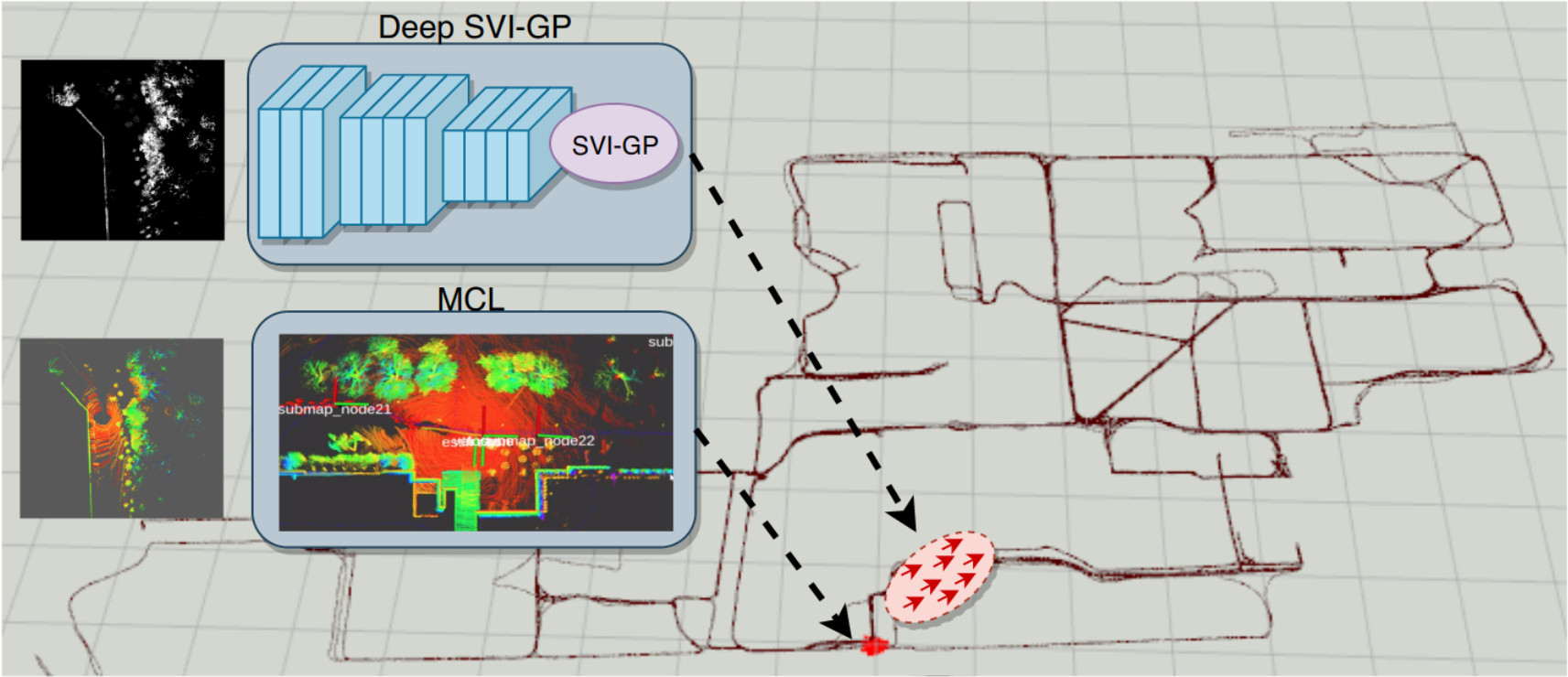}
  \includegraphics[width=\columnwidth]{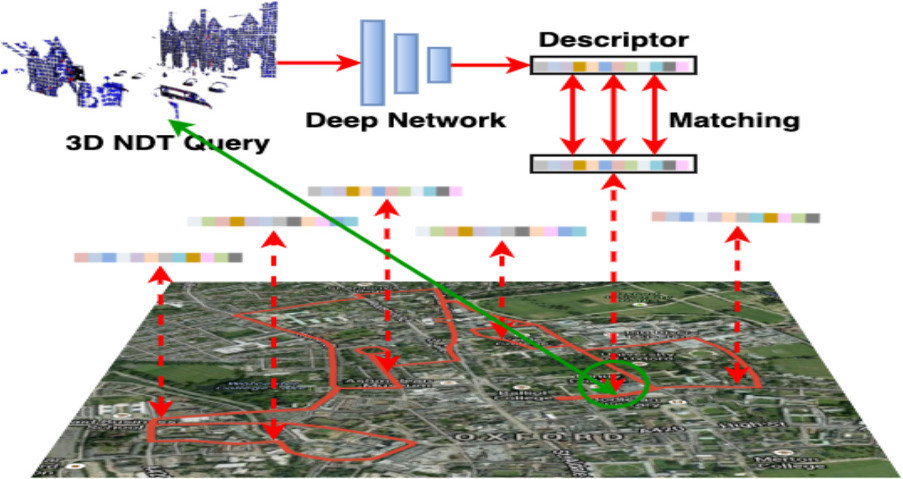}
  \caption{Examples of metric-based (upper) \cite{ls20icra} and one-shot global (lower) \cite{ls21icra} robot localization based on 3D LiDAR.}
  \label{fig:robot_localization}
\end{figure}

The approaches of robot localization using LiDARs can be categorized as metric-based methods and one-shot global localization methods (see Fig.~\ref{fig:robot_localization}).
The former aims to obtain a 3D geometrical map and re-localize the robot with explicit 3DOF or 6DOF global poses.
Whereas in the latter, the geometrical world is discretized as appearance locations (also known as places) in which the robot localization can be inferred by global pose regression or place retrieval.
Both categories can work standalone according to the application scenarios, but noting that they can also be integrated as a hybrid method such as using global localization to initialize the metric localization to conduct a coarse-to-fine localization.
Below we investigate the two types of methods separately.

\subsubsection{Metric-based localization}

A classic way of LiDAR metric localization is first to use the Simultaneous Localization And Mapping (SLAM) method to build a geometrical map as a priori and then employ the Monte Carlo Localization (MCL) to obtain accurate and persistent localization.
MCL is robust in principle, and by using a multi-modal belief distribution, global localization can also be performed through a very uncertain initial estimate of the robot position.
However, naive initialization using uniform particle distribution does not scale well to a map of realistic size.
Some authors~\cite{ryde20103d} alleviate this problem by working with a multi-resolution grid over the map, which makes it possible to scale to slightly larger maps.
To avoid uniform distribution problems caused by scaling, we recently proposed to learn the appearance of the map~\cite{ls20icra}, which leverages learning-based localization and filtering-based localization, to localize the robot efficiently and precisely through seeding MCL with a deep-learned distribution.

Another strategy is to design a distribution from observations and sample particles from that~\cite{thrun_monte_2000,he2013observation,kucner-2015-mcl}.
Another possibility is to use external information, such as WiFi beacons~\cite{seow2017detecting} or a separate map layer that governs the prior probability of finding robots in each part of the map~\cite{oh_map-based_2004}.
Our recent work~\cite{ls20icra} proposed to directly process point cloud data, and showed the superiority of robot localization on a map of a large environment of approximately $0.5~km^2$.

\subsubsection{One-shot global localization}

The state-of-the-art one-shot global localization methods can be categorised into three main streams, including regression-based, intermediate-representation-based and global-feature-based.
Following the success of deep pose estimation in image-based global localization, some methods~\cite{wang2020pointloc,akai2020hybrid} propose to use deep regression networks to learn the 6DOF global pose.
These regression-based approaches are extremely efficient but not generalizable.
In other words, a new model is required to be trained for new environments.

Intermediate-representation-based methods~\cite{dube2017segmatch,segmap2018,tinchev2019learning,kong2020semantic} first segment the LiDAR map into intermediate representations (i.e. semantic segments) then register the segmented parts, rather than directly registering the point clouds.
The intermediate parts can be described as distinctive features~\cite{dube2017segmatch} or deep-learned features~\cite{segmap2018,tinchev2019learning, kong2020semantic} may be used to eliminate the false positives.
Though the matching efficiency can be improved by using intermediate representations, significant running time is spent on cloud segmentation and real-time performance cannot be guaranteed.
Therefore, these methods are mostly used for loop-closure detection rather than large-scale localization.

The mainstream methods are based on global features, where a generalizable feature extractor is designed or learned to obtain the place signature for retrieval.
Before deep learning began dominating the machine learning community, hand-designed heuristics features such as M2DP~\cite{he2016m2dp}, Scan Context~\cite{kim2018scan} and DELIGHT~\cite{cop2018delight} were well-studied to represent the 3D point cloud for the localization.
In terms of learning-based methods, the Scan Context Image-based network~\cite{kim20191} and OverlapNet~\cite{chen2020overlapnet} convert the 3D LiDAR scan to a 2D image according to geometric knowledge, then deploy a 2D convolution-like network to learn a representation.

Instead of using hand-crafted features, PointNetVLAD~\cite{angelina2018pointnetvlad} combines PointNet~\cite{qi2017pointnet} and NetVLAD~\cite{Arandjelovic_2016_CVPR} to learn a global descriptor based on metric learning. 
However, the PointNet-like architecture ignores the spatial distribution and contextual cues within the 3D point cloud. 
Extracting efficient contextual information from the irregular 3D point cloud is another important challenge for 3D loop-closure detection. 
The following work LPD-Net~\cite{liu2019lpd} employs a classical DGCNN~\cite{wang2019dynamic}-like network to enhance the feature descriptor by KNN-based aggregation in both feature space and Cartesian space.
Our recent work~\cite{ls21icra} proposed to employ a 3D Normal Distribution Transform (NDT) representation to condense the raw, dense 3D point cloud as probabilistic distributions (NDT cells) to provide the geometrical shape description, so as to achieve real-time and large-scale place recognition using 3D point clouds.

\subsection{Long-term Autonomy}
\label{sec:longterm_autonomy}

Advances in 3D LiDAR-based mapping and localisation enabled reliable operation of mobile robots in large-scale environments, leading to their deployments in agriculture~\cite{agricultural}, logistical~\cite{ls21iros} or search-and-rescue operations~\cite{DARPA1,DARPA2}.
The ability of reliable operation, however, brought another problem: the need for 3D (or 4D) maps that would reflect the fact that the world is perpetually changing.
While several works showed that taking into account environment dynamics when creating and updating 2D LiDAR-based maps improves localisation~\cite{biber,tipaldi,krajnik2016persistent}, full 3D map update based on a 3D LiDAR is computationally intensive, which requires more advanced map management techniques~\cite{pomerleau2014long}.
In particular~\cite{pomerleau2014long} shows that long-term 3D LiDAR observations can be used not only to update the scene geometry, but also to segment dynamic objects like pedestrians or cars.
The information about dynamic objects can be further processed to identify dominant motion flows~\cite{kucnerconditional,ls18icra,vintr19ecmr}, which can be used to improve robot navigation~\cite{vintr20iros,cliff} in crowded environments.
Apart from detecting the change~\cite{ayoung}, long-term, 3D observations allow to learn the temporal patterns of the changes and use this knowledge to forecast the environmental states~\cite{ls18ral,fremen}, which improves the performance of mobile robots in the long-term.
A comprehensive review of the maps that leverage the long-term observations to characterise the environment dynamics is provided in~\cite{tomek}.

\subsection{Adverse Weather Conditions}
\label{sec:adverse_conditions}

When robots equipped with 3D LiDAR are working outdoors, different weather conditions, especially bad weather, are inevitable.
In addition to using other types of ``weather robust'' sensors, such as radar, to ensure the quality of perception, the processing of the 3D LiDAR data itself is also important due to its irreplaceable accuracy of ranging.
To this end, researchers first hope to better understand the impact of various adverse weather on the sensor and evaluate it, then model this impact, usually in the form of noise, and finally think of ways to mitigate or even remove the adverse effects, that is, denoising.

\subsubsection{LiDAR performance evaluation}

Peynot \emph{et al.}~\cite{Peynot2009} observed that dust particles can be detected by laser sensors and may hide obstacles behind the dust cloud.
Michaud \emph{et al.}~\cite{michaud2015towards} presented behavior analysis of several LiDARs in snow. The authors concluded that the falling snow mainly generates noisy points in a range of 10 m from the LiDAR.
Hasirlioglu \emph{et al.}~\cite{Hasirlioglu2017} investigated effects of exhaust gases at low temperatures on sensor quality and found that exhaust gases are visible for laser scanners and lead to degraded performance.
Ashraf \emph{et al.}~\cite{Ashraf2018} used Velodyne HDL-64E at $905nm$ wavelength to analyze fog attenuation, and applied natural neighbor interpolation to construct 2D images from attenuated 3D LiDAR data.
They found that the introduction of a fog attenuation model would lower contrast and less sharp 2D images, making object detection more difficult.
Filgueira \emph{et al.}~\cite{Filgueira2017} carried out real-world measurements with a Velodyne VLP-16 LiDAR for a couple of months to evaluate the influences of different rainfall levels on different reflective materials.
They found that the returned intensity and the number of detected points decrease with the increase of rain in all targets under study, and the highest performance decrease occurs on the asphalt pavement.
In contrast, Goodin \emph{et al.}~\cite{goodin2019predicting} claimed that heavy rain doesn't have significant influence on LiDAR's performance of obstacle detection by running the the Mississippi State University Autonomous Vehicle Simulator (MAVS) with a developed mathematical model for the LiDAR performance in the rain.
Yoneda \emph{et al.}~\cite{yoneda2019automated} compared the performance of multiple sensors including LiDAR, millimeter wave radar, and camera by indoor and outdoor testing and indicated that it's important to estimate the degree of adverse weather conditions to perform practical operations.

\begin{figure*}[t]
  \centering
  \includegraphics[width=\textwidth]{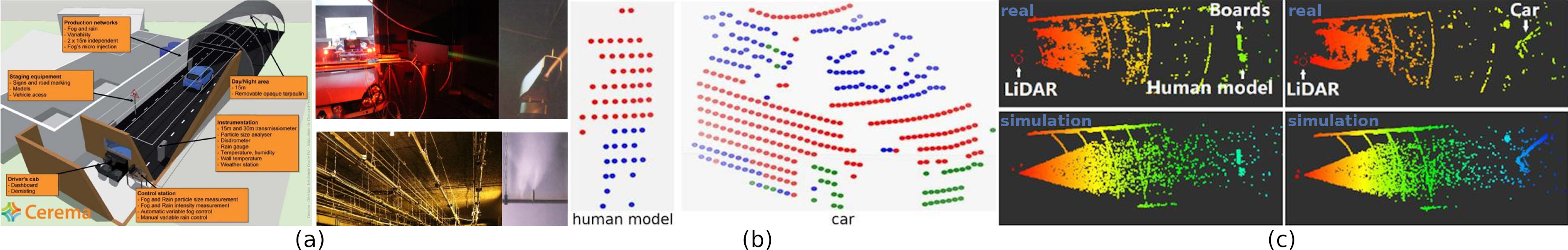}
  \caption{(a) In the Cerema climate chamber, various level of artificial fog and rain can be generated and quantified by a transmissometer for meteorological visibility measurement~\cite{you20fog}.
    (b) The point clouds of human dummy and vehicle captured by a velodyne VPL-32 LiDAR in Cerema's chamber~\cite{yang20iros}.
    (c) The real measurements and simulation results of LiDAR point cloud under fog (bird's eye view)~\cite{yang20iros}.}
  \label{fig:cerema-lanoising}
\end{figure*}

Kutila \emph{et al.}~\cite{Kutila2018} compared the performance of LiDARs with different wavelengths of $905nm$ and $1550nm$ under different fog levels, and further indicated that there is no reason to not use $1550nm$ wavelength, which due to eye safety regulations gives an opportunity to use 20 times more power compared to the traditional $905nm$.
Jokela \emph{et al.}~\cite{jokela2019testing} investigated the performance of various LiDARs from the manufacturer of Ibeo, Velodyne, Ouster, Robosense, and Cepton in a climate chamber build by Cerema (see Fig.~\ref{fig:cerema-lanoising}), a French public agency, and further tested in outdoor environment under turbulent snow conditions.
They declared that all the tested sensors' performance decreased in harsh weather, and it is difficult to say which one is better.
Karl \emph{et al.}~\cite{Karl2021} analysed the intensity and the number of points detected on a calibrated target and in the frustum between the sensor and the target in artificial fog and rain with the LiDARs of Velodyne, Ouster, Livox, Cepton, and AEye in different echo modes.
They found that the $1550nm$ AEye LiDAR allows better quality of point clouds, and a non repetitive scan pattern from the Livox LiDAR shows good performances under fog.
Meanwhile, the multi-echo modes result to more information and noise under adverse weather conditions.
Recently, Alexander \emph{et al.}~\cite{Alexander2020} performed a more comprehensive comparison of ten different LiDARs with data collected in both a similar climate chamber and dynamic traffic from a vehicle driven on urban roads.

\subsubsection{Noise modeling}

After knowing the impact of various weather on the performance of LiDAR, it is necessary to model this impact.
This can be approached on two ways: bottom-up (physical model) and top-down (data-driven).
For the former, Rasshofer \emph{et al.}~\cite{Rasshofer2011} mathematically modeled the LiDAR signal transmit pulse and spatial impulse response, and derived the spatial impulse response of hard targets.
For the soft targets, the spatial impulse response function of rain or fog was described by the backscattering coefficient, and the authors gave a rough estimation of the extinction coefficient in case of dry and wet snow.
Roy \emph{et al.}~\cite{roy2020physical} modeled the interaction of snowflakes with 3D LiDAR taking into account the physical model of laser power transmission and the statistical model of snowflakes' characteristics.  
Hahner \emph{et al.}~\cite{hahner2021fog} developed a LiDAR noise simulator under fog according to the physical equation of laser power transmission.
However, since LiDAR manufacturers usually use undiscloseable internal mechanisms to output the received laser intensity for commercial reasons, the authors had to assume that the intensity is a linear function of the maximum received laser power.

In view of the current difficulties in accurate physical modeling of LiDAR, the top-down approach seems more promising, and for this, data is the key.
Embarrassingly, collecting and annotating data under real adverse weather conditions suffer from accurate weather forecasting and quantification, as well as cost-intensive post-data processing and labeling.
Therefore, to some extent, the emulation of adverse weather, such as the aforementioned Cerema chamber, would help the modeling of data noise.
Bijelic \emph{et al.}~\cite{Bijelic2018} tested four state-of-the-art LiDARs from Ibeo and Velodyne in controlled conditions in the Cerema chamber under fog, and presented disturbance patterns for these four LiDAR systems in fog.
In the same chamber, our previous work investigated the performance of a typical $903nm$ ToF LiDAR under fog.
We first trained a machine learning model to predict the minimum fog visibility that allows this type of LiDAR to return to the correct range~\cite{you20fog}, and then proposed a two-stage method to model the relations between the ranging performance of a ToF LiDAR and fog levels~\cite{yang20iros,yang21tits}.
Novelly, giving certain fog visibility values, a Gaussian process Regression (GPR) model was first trained to predict whether a laser can successfully output a true detection range or not.
If not, a Mixture Density Network (MDN) then provided a probability prediction of the noisy measurement range.
Moreover, compared with the bottom-up modeling of LiDAR performance under fog based on physical models, the above top-down methods by machine learning can directly utilize the raw data of LiDAR under clear weather conditions as the model input.

\subsubsection{Denoising}

LiDAR denoising mainly adopts filtering methods.
Some existing works focus on reducing the random and unwanted variations of LiDAR signals in order to obtain the significance of the signal as much as possible, such as wavelet filter~\cite{Yin2006}, mathematical morphology method~\cite{Sun2017}, and Singular Value Decomposition (SVD) approach~\cite{Azadbakht2013}.
However, as mentioned earlier, the signal processing unit of an off-the-shelf LiDAR is usually regarded as a black box.
Therefore, more research focuses on removing the noise points in the point cloud data generated by the sensor.

Nie \emph{et al.}~\cite{Nie2011} proposed an algorithm to classify scattered points according to the surface variation in order to detect the outliers.
Cao \emph{et al.}~\cite{Cao2013} proposed to denoise LiDAR point cloud based on feature selection of bilateral filtering, which requires, however, a longer period of time to determine the feature selection.
Zhu \emph{et al.}~\cite{Zhu2015} distinguished noise points using the local density.
Javaheri \emph{et al.}~\cite{Javaheri2017} indicated that graph-based denoising algorithms can significantly improve the point cloud quality, and objective metrics that model the underlying point cloud surface can correlate better with human perception.
Shan \emph{et al.}~\cite{Yonghua2017DenoisingAO} processed the point cloud in 3D grid and filtered out the noise points according to the spatial neighborhood relationship.
The Point Cloud Library (PCL)~\cite{Holz2015} contains a module for point cloud data outlier (noise) removal, which provides a easy-to-use baseline for denoising of LiDAR data.
The implementation of outlier removal in PCL is based on the statistical analysis of LiDAR points and their neighbors in the input data.

Judging from the results published so far, there is not much research on methods for denoising LiDAR point clouds under fog, rain or snow conditions.
Charron \emph{et al.}~\cite{8575761} proposed a method to eliminate snow noise by processing point clouds using a 3D outlier detection algorithm.
Heinzler \emph{et al.}~\cite{heinzler2020cnn} regarded the point cloud denoising as a semantic segmentation problem, and applies a U-Net architecture based Convolutional Neural Network (CNN) to the point cloud data recorded in the Cerema climate chamber for training and validation.
Duan \emph{et al.}~\cite{duan2021low} first generated 2D data from the point cloud through Principal Component Analysis (PCA), and then reduced noise through an adaptive clustering method. But they didn't really test on point clouds in adverse weather.
JI-IL \emph{et al.}~\cite{park2020fast} removed snow particles based on the intensity between laser points to improve speed and accuracy of existing distance-based filters.
Kurup \emph{et al.}~\cite{kurup2021dsor} proposed a PCL-based filter for snow conditions by calculating the outlier threshold with k-nearest neighbor searching.

\subsubsection{Datasets}

Some well-known datasets, such as KITTI~\cite{KITTI}, Waymo~\cite{Waymo}, and nuScenes~\cite{nuScenes}, have played a recognized role in promoting the development of outdoor robotics and autonomous driving. 
Although these datasets mainly focus on various problems in good weather conditions, they are completely meaningful in the early stages of technological development.
In recent years, with the rapid development of relevant technologies, the demand for all-weather operation of unmanned ground vehicles has arisen accordingly.
So far, several datasets containing LiDAR point clouds have been released, which deliberately collect sensor data in adverse weather, aiming to promote robot autonomy in edge conditions such as fog, rain, and snow.

Gruber \emph{et al.}~\cite{gruber2019pixel, Gruber_2019_ICCV} released data collected in the Cerema chamber and provided a pixel-level depth map as ground-truth.
Another dataset, the See Through Fog (STF) dataset~\cite{Bijelic20}, was acquired by driving over $10,000 km$ in northern Europe across different kinds and levels of adverse weather, with $100k$ labels for object detection.
LiDAR Benchmarking and Reference (LIBRE) dataset~\cite{Alexander2020} focuses on evaluation of ten different LiDARs in both indoor and outdoor environments.
Canadian Adverse Driving Conditions (CADC) dataset~\cite{cadcd} was collected specifically for snow conditions, with a Velodyne VLP-32 LiDAR, eight RGB cameras, and a GNSS-RTK/IMU pair.
The labels provided contain not only the 2D/3D object bounding boxes, but also the object trajectories.
SMART-Rain dataset~\cite{zhang21iros} was collected under different rainfall levels, in which the LiDAR data under clear weather conditions was used as reference for rainfall intensity and LiDAR data qualification estimations.
In addition to the LiDAR and stereo camera, the RAdar Dataset In Adverse weaThEr (RADIATE)~\cite{sheeny2020radiate} provides high-resolution radar images with 2D/3D object labels.
Our previous work in this regard includes the EU long-term dataset~\cite{yz20iros}, that was collected with four LiDARs, a front-facing and a rear-facing stereo camera, two side-pointing fisheye cameras, a long range radar, a GNSS-RTK and an IMU, across different seasons and weather conditions.
Winter Adverse Driving dataSet (WADS)~\cite{kurup2021dsor} was collected in the snow belt region in Michigan, USA. Point-wise segmentation labels of 22 classes were provided, including the noises of point clouds in severe winter weather.
A comparison of the above datasets is detailed in Table~\ref{tab:datasets}.

\begin{table*}[t]
  \caption{LiDAR datasets for autonomous driving under adverse weather conditions}
  \label{tab:datasets}
  \begin{center}
    \begin{tabular}{|c|c|c|c|c|c|}
      \hline
      \textbf{Dataset} & \textbf{LiDAR sensors} & \textbf{Ground-truth} & \textbf{Location} & \textbf{Weather} & \textbf{Other sensors}\\
      \hline\hline
      DENSE/STF~\cite{Bijelic20} & \makecell[c]{$1\times$ Velodyne HDL64-S3 LiDAR \\ $1\times$ Velodyne VLP-32 LiDAR} & \makecell[c]{Indoor: \\ Depth estimation \\ Outdoor: \\ Object detection} & \makecell[c]{Indoor \& \\ Outdoor: \\ Northern Europe} & \makecell[c]{Clear \\ Rain \\ Fog \\ Snow} & \makecell[c]{$1\times$ Stereo camera \\ $1\times$ Gated camera \\ $1\times$ FIR camera \\ $1\times$ Front radar} \\ 
      \hline
      LIBRE~\cite{Alexander2020} & \makecell[c]{$5$ different Velodyne LiDARs \\ $2$ different Hesai LiDARs \\ $2$ different Ouster LiDARs \\ $1 \times$ RoboSense RS-32 LiDAR} & \makecell[c]{Indoor: \\ Object distance \\ Outdoor: \\ Object detection \\ Self-localization} &  \makecell[c]{Indoor \& \\ Outdoor: \\ Japan} & \makecell[c]{Clear \\ Rain \\ Fog \\ Strong light} & \makecell[c]{$1\times$ RGB camera \\ $1\times$ IR camera \\ $1\times$ $360^{\circ}$ camera \\ $1\times$ Event camera} \\ 
      \hline
      CADC~\cite{cadcd} & $1\times$ Velodyne VLP-32C LiDAR & \makecell[c]{Object detection \\ Object tracking \\ Self-localization} & \makecell[c]{Outdoor: \\ Canada} & \makecell[c]{Clear \\ Snow} & \makecell[c]{$8\times$ RGB camera} \\ 
      \hline
      RADIATE~\cite{sheeny2020radiate} & \makecell[c]{$1\times$ Velodyne HDL-32e LiDAR} & \makecell[c]{Object detection \\ Self-localization} & \makecell[c]{Outdoor: \\ UK} & \makecell[c]{Clear, Rain \\ Fog, Snow} &\makecell[c]{ $1\times$ Stereo camera \\ $1\times$ $360^{\circ}$ Radar} \\ 
      \hline
      SMART-Rain~\cite{zhang21iros} & $1\times$ RoboSense RS-32 LiDAR & Rainfall intensity & \makecell[c]{Outdoor: \\ Singapore} & \makecell[c]{Clear \\ Rain} & \makecell[c]{ $1\times$ RGB camera \\ $1\times$ Rainfall sensor }\\
      \hline
      EU long-term~\cite{yz20iros} & \makecell[c]{$2\times$ Velodyne HDL-32E LiDAR \\ 1$\times$ Ibeo LUX 4L LiDAR \\ 1$\times$ SICK LMS100 2D LiDAR} & Self-localization & \makecell[c]{Outdoor: \\ France} & \makecell[c]{Clear \\ Snow} & \makecell[c]{$2\times$ Stereo camera \\ $2\times$ Fish-eye camera \\ $1\times$ Front radar} \\ 
      \hline
      WADS~\cite{kurup2021dsor} & \makecell[c]{3 different 64-layer LiDARs \\ 2$\times$ Velodyne VLP-16 LiDAR} & Semantic segmentation & \makecell[c]{Outdoor: \\ USA} & \makecell[c]{Snow} & \makecell[c]{$1\times$ RGB camera \\ $1\times$ NIR fish-eye camera \\ $1\times$ LWIR camera} \\ 
      \hline
    \end{tabular}
  \end{center}
\end{table*}

\subsection{Sensor fusion}

Because of its physical characteristics, different types of sensors have their own limitations and edge cases.
Typically, camera can reliably detect objects as it can provide color and texture information, but its detection performance is limited by the visual scope and affected by the lighting conditions, and it also suffers from the accurate localization of objects in space.
LiDAR can provide up to 360 degrees of horizontal FoV and a measurement distance of tens to hundreds of meters, but the measurement returns are discrete points in space, making objects difficult to identify due to lacking easy-to-interpret features such as color and texture.
Hence, compared with robot perception that relies on a mono-modal sensor solution, academia and industry agree that using sensors with multiple modalities is more promising.
An obvious benefit is that redundant cross-sensory data can provide more comprehensive environmental information.

\begin{figure}[t]
  \centering
  \includegraphics[width=\columnwidth]{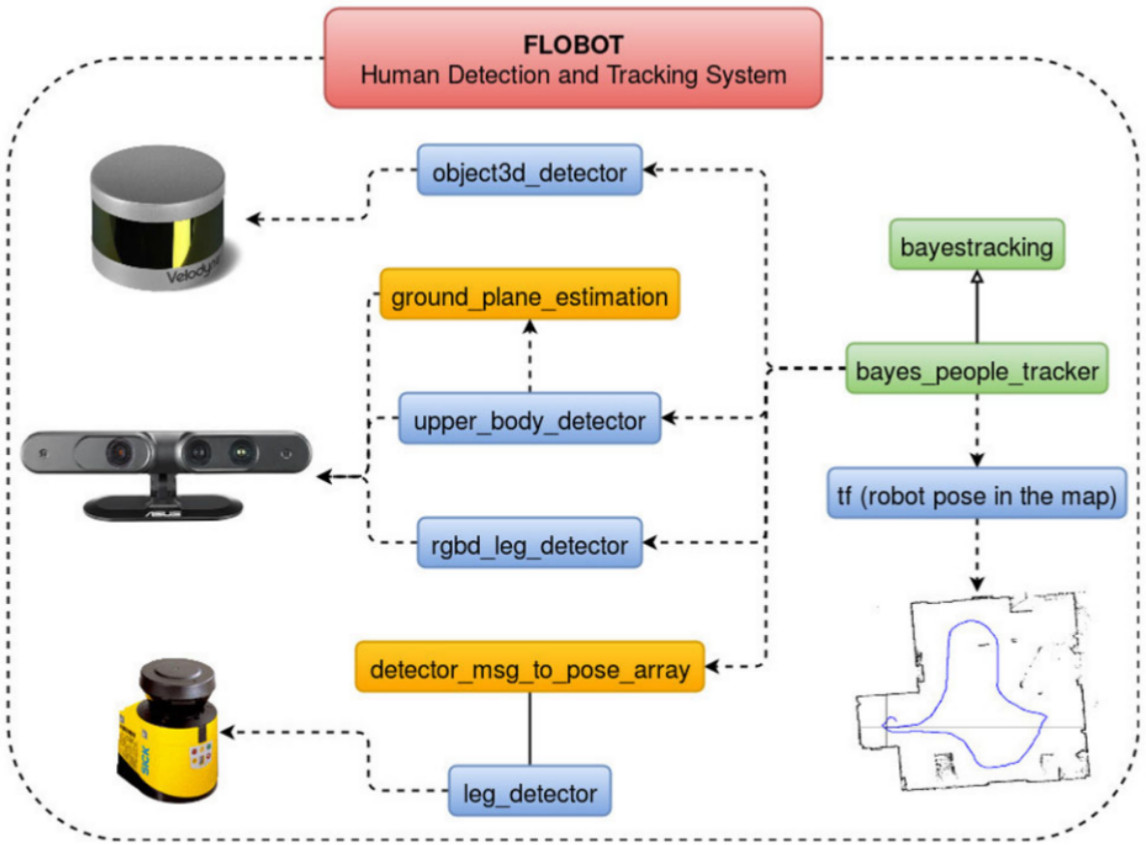}
  \includegraphics[width=\columnwidth]{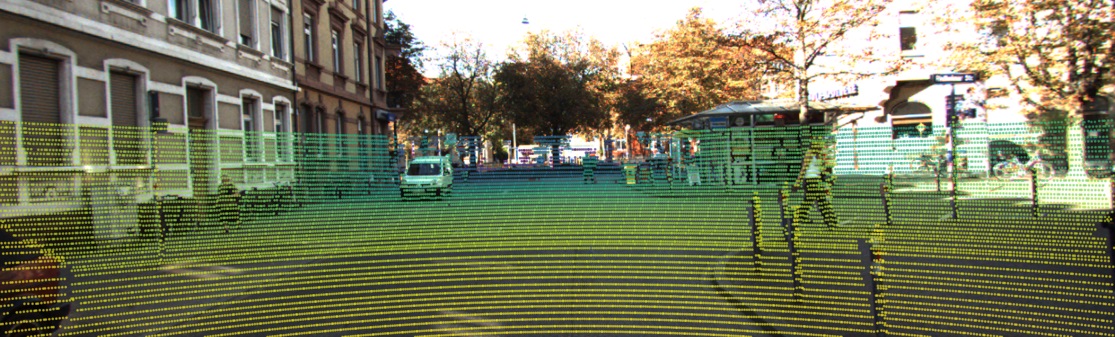}
  \caption{Examples of high-level (upper) \cite{zhimon20jist} and low-level (lower) sensor fusion.}
  \label{fig:sensor_fusion}
\end{figure}

Regardless of the single modal or multiple modals of the sensors, the existing methods for sensor fusion can be roughly divided into two categories, including high-level fusion and low-level fusion (see Fig.~\ref{fig:sensor_fusion}).
The former mainly refers to firstly extracting interesting information independently from each sensor's data, and then realizing fusion through information association.
Typical examples include the multisensor-based tracking-by-detection pipeline~\cite{zhimon20jist}, and multisensor-based environment exploration~\cite{DARPA1}.
The latter refers to the direct matching of sensory data, such as the concatenation of point clouds from different LiDARs~\cite{sualeh19sensor}, or the point-pixel registration~\cite{KITTI}.

If we look at the combination of different types of sensors, camera with 3D LiDAR is probably the fastest-rising combination in literature.
\cite{premebida14iros} trained a deformable parts detector for pedestrians using different configurations of optical images and their associated 3D point clouds, individually and in conjunction, and showed that the combined use results in detection performance exceeding that of the RGB-only system.
\cite{gonzalez15iv} explored the fusion of the depth map obtained by high-definition 3D LiDAR and the RGB map obtained by camera, and provided a broad assessment of how pedestrian detection performance is affected by the use of sensors alone and in combination.
\cite{iliad} conducted a systematic cross-modal analysis of sensor-algorithm combinations typically used in robotics, and compared the performance of state-of-the-art person detectors for 2D range data, 3D LiDAR and RGB-D data as well as selected combinations thereof in a challenging industrial use-case.
Our previous work~\cite{zhimon20jist} also proposed the combined use of a RGB-D camera, a 2D and a 3D LiDAR for a professional cleaning robot to robustly detect and track people in its working environment.

More interestingly, as mentioned in Sec.~\ref{sec:adverse_conditions}, in fog, rain, and snow, small droplets and snowflakes cause the distortions of LiDAR depth estimation, while there is limited impact on radar sensors.
To address this challenge, Majer \emph{et al.}~\cite{majer19ecmr} proposed an online method which uses the LiDAR-based static detector provides labels for the radar point clouds to train a SVM classifier as a dynamic detector.
During adverse weather conditions, a learning-based sensor fusion scheme was designed to make the system more rely on the radar to achieve higher accuracy of pedestrian detection and localization.
George \emph{et al.}~\cite{broughton2020learning} further involved PointNet~\cite{qi2017pointnet} in the online learning framework of sensor fusion to benefit from the current progress of deep learning methods of point cloud processing.
PointNet was used to add a class label to each point of radar data, which was classified as clusters and finally sent to the sensor fusion module.
Besides fusion after independently processing the different sensor streams,
Bijelic \emph{et al.}~\cite{Bijelic20} performed a end-to-end mechanism for adaptive fusion of LiDAR, RGB camera, gated camera, and radar sensor streams in adverse weather, in which all sensory data was projected into the camera 2D coordinate and an entropy map for each sensor was calculated to extract and fuse the effective features from different layer feature maps.

Furthermore, given the capabilities of sensors are different, the transfer of knowledge between different types of sensors, or say, one category learns from another, autonomously, has become a constructive proposal.
It can undoubtedly save humans from tedious offline training tasks, and in mobile robotics, this learning paradigm further enables robots to learn on-site in their deployment place to adapt to changes in the environment.
This paradigm has been used not only for LiDAR~\cite{yz18iros,yang21itsc}, but also for millimeter wave radar with similar data representation~\cite{majer19ecmr,broughton2020learning}, not only for indoor service robots~\cite{yz18iros}, but also for autonomous driving perception in urban environments~\cite{yang21itsc}.

\section{Applications}
\label{sec:applications}

3D LiDAR has been adopted by a growing number of researchers and industries, thanks to its ability to provide accurate geometrical information (i.e. point cloud) about its environment over a long range and wide angle.
At the earliest, the Velodyne's 64-layer 3D LiDAR contributed to the success of Stanford's Junior robot car~\cite{montemerlo08jfr} in the 2007 DARPA Grand Challenge, but in the meantime forged its own attainment.
Junior's primary sensor for obstacle detection such as pedestrians, signposts, and cars, is the 3D LiDAR.
The success of Junior also directly affects and promotes the development of 3D LiDAR in autonomous driving~\cite{you20spm,yz20iros,KITTI,oxford-dataset,KAIST,cadcd,nuScenes,yang21itsc,you20fog,yang20iros,Waymo,yang21tits}.
On the other hand, the demand for unconventional sensors typified by 3D LiDAR is also present in the field of service robots~\cite{yz17iros,yz19auro,yz18iros,ls20icra,DARPA2,broughton2020learning,vintr20iros,zhimon20jist,vintr19ecmr,vintr19icra,ls18icra,ls18ral}.
Furthermore, in the face of the recent severe worldwide public health crisis, namely COVID-19, 3D LiDAR has also shown encouraging results and broad application prospects.
In the remainder of this section, we will discuss related application areas by category.

\subsection{Service Robots}

3D LiDARs are currently mainly equipped with service robots for professional use that usually work in indoor public~\cite{zhimon20jist} and outdoor environments~\cite{DARPA1,ls21iros,agricultural}, while relatively rare for personal robots that are typically deployed in small scenes such as domestic environments~\cite{hsr,STRANDS,ENRICHME}.
It is generally installed on the top of the robot to have a good FoV (e.g. less occlusion) and can be used for mapping, localization, object detection and tracking, semantic segmentation, scenes understanding, and so forth.
For example, a robotic scrubber-dryer (floor washer with brushes and suction to dry the floor, as shown in Fig.~\ref{fig:LiDAR-robot-car}(b)) is developed for large public premises such as supermarkets, airports, trade fairs and hospitals, in which a multisensor perception system is employed including a 3D LiDAR to help long-range and wide-angle human detection and tracking for safety, which is highly demanded for indoor space with dense public moving around.
Another example is an autonomous forklift developed for warehouse logistics~\cite{iliad}, in which a 3D LiDAR is used for mapping with moving obstacles removal (i.e. human detection), and a 2D obstacle grid map is then extracted from the fully-3D map to be used for further navigation and motion planning.
For outdoor cases, 3D LiDARs are widely used for last-mile delivery~\cite{ls21iros}, exploration and search and rescue~\cite{DARPA2}, agricultural robots~\cite{agricultural} and more.

\subsection{Autonomous Driving}

Although Tesla does not want to admit it, 3D LiDAR has become a \emph{de-facto} standard for driverless cars~\cite{safety}.
In order to obtain a richer representation of the environment (i.e. more laser echoes) and also minimize visual blind spots, vehicles are usually equipped with multiple LiDARs to be used in conjunction~\cite{yz20iros,KAIST,Waymo,sualeh19sensor}.
In addition to using the sensor for robotic tasks such as object detection and tracking, it is also used for the production of high-definition maps\footnote{It is worth pointing out that the high-resolution map should also contain more road information such as lanes, traffic signs, traffic rules, and more.} - a key module of autonomous driving - and localization based on it.
However, due to its physical characteristics, the ToF LiDAR is affected by adverse weather conditions as mentioned before, in particular due to water droplets or dust in the atmosphere.
Indeed, relying solely on 3D LiDAR is not enough, which is why other modalities like cameras and radars are also integrated into today's vehicle perception systems.

\subsection{Public Health Crises}

At present, in the face of a global pandemic, various technical equipment from all walks of life are mobilized, including the use of 3D ToF LiDAR.
The most direct application is for human detection and tracking.
The technology can effectively monitor human social distance while respecting personal privacy~\cite{sathyamoorthy2020covid}.
Recent work can even show whether a person wears a mask~\cite{loey2021hybrid}.
With the use of other anti-epidemic equipment such as thermal sensors, people can also identify the infected person and contacts in time to make corresponding treatments.
Chen \emph{et al.}~\cite{chen2020overlapnet} developed a surveillance robot with multiple cameras and a 3D LiDAR mounted on the quadruped platform to detect and track nearby pedestrians for promoting social distancing using  crowd-aware navigation algorithm.
Besides keeping social distance and monitor body temperature of pedestrians, service robots are mainly used for disinfection and delivery.
In order to reduce the risk of cross-infection, disinfection robots replace manual labors and perform disinfection work in hospitals~\cite{mcginn2021exploring}.
Delivery robots distribute food and medicine automatically to reduce the direct contact between hospital staff and patients. In addition, the distribution robots also play an efficient role in the material warehouse~\cite{chen2020containing}. \tao{As indicated by Yang \emph{et al.}~\cite{yang2020combating}, China explores actively the applications of robots against Covid-19. Furthermore, developing robots with capability of social interactions to alleviate the mental health impact caused by quarantine is still a challenge.}

\section{Perspectives}
\label{sec:perspectives}

Based on our survey and research experience, we are generally optimistic about the application prospects of 3D ToF LiDAR in mobile robotics, although this is still accompanied by many foreseeable challenges.
The first reality is that the performance and price of the current mass-produced LiDAR still show a non-linear proportionality.
This reflects from one aspect the importance of datasets to promote the development of related technologies.
Fortunately, there are many currently available in community.
In the future, we hope that more high-quality datasets of different scenarios will appear, especially based on multisensor systems.

As we mentioned earlier, ToF LiDAR has its own shortcomings including the sparse point cloud generated and the difficulty of interpreting data.
To make up for these, the former mainly depends on the development of hardware, while the latter has be used in conjunction with other modal sensors.
From the perspective of object detection, the former mainly affects the result of point cloud segmentation, while the latter is not conducive to object recognition.
In addition, from an algorithm perspective, similar to those based on visual sensors, the challenges of object detection are mainly in the segmentation of objects that are too close, medium and long distances, occlusion, truncation, and adverse weather, especially for pedestrians.
Responding to these challenges usually requires additional reasoning mechanisms, which can rely on semantic segmentation of the scene, multi-target tracking, spatiotemporal modeling, point cloud denosing, and so on.

LiDAR-based mapping, odometry, and localization show a significant accuracy beyond other modal sensors.
From our research perspective, more attention needs to be paid to the generality and scalability of the robot localization systems.
The motivations mainly include the insufficient generalization ability of current regression-based methods and the high demand on computational resources of point cloud-based place retrieval. 
Moreover, from a technical level, the core challenge of large-scale metric localization is the initialisation of particles in MCL.
For this purpose, the global pointing system (GPS) can be used to initialize the particles.
While in GPS-denied environments, other alternatives such as global pose estimation or appearance descriptors based place recognition can be employed to provide one-shot localization.

As Operational Domain Designs (ODDs) is one of the main concerns in today's autonomous driving industries, the adaption to different weather is the key to specify performance boundaries of the system. At present, the traditional de-noising methods of LiDAR in adverse weather still need to be improved in both accuracy and speed, and the generalization of deep learning method needs to be further verified. Since annotation of point cloud is difficult in such edge conditions, LiDAR performance simulator becomes an efficient tool to generate training and validation data. Meanwhile, it's also important to estimate the weather conditions using the current sensors to decide whether the autonomy system should be activated.

In terms of application, we are particularly optimistic about the last-mile delivery, as a field that connects indoor robots and autonomous driving, is an excellent testbed for integrating scenario-specific technologies from both sides.
Especially during current public health crisis, the food and goods delivery still relies on manual labors.
In the future, it is foreseeable that mobile robots, including self-driving cars, will become part of our daily lives.
3D ToF LiDAR is expected to play a key role in all these existing or potential applications to contribute the precise environment perception and self-localization, long-term autonomy, and all-weather operating capabilities.

\section{Conclusions}
\label{sec:conclusions}

In this review, we systematically introduced the use of 3D ToF LiDAR in mobile robotics.
We first briefly introduced its ranging principle, physical structure and characteristics, as well as data representation and processing.
We then listed several main research axes based on our research focuses including object detection, robot localization, long-term autonomy, adverse weather conditions, and sensor fusion, followed by some of its application areas including service robots, autonomous driving, and especially in responding to the ongoing global pandemic, and finally illustrated some prospects for future research.

It can be seen from our review that 3D ToF LiDAR has been widely studied and applied in our community, especially for large indoor and outdoor environments.
Based on its bright development potential and large-scale application prospects, we hope that our efforts can promote the development of related technologies, especially to provide effective guidance for the rapid deployment of existing methods.
In addition, we should also pay attention to the relevance of the aforementioned research axes.
For example, robust static and dynamic object detection is conducive to the mapping and localization of the robot, and both, together with the consideration of adverse weather conditions, are conducive to the long-term robot autonomy.
Finally, we invite the community to make comparisons following the benchmarking we provided, and even contribute code to open source libraries when possible.

\bibliographystyle{IEEEtran}
\bibliography{ref.bib}

\begin{thebibliography}{100}
\providecommand{\url}[1]{#1}
\csname url@samestyle\endcsname
\providecommand{\newblock}{\relax}
\providecommand{\bibinfo}[2]{#2}
\providecommand{\BIBentrySTDinterwordspacing}{\spaceskip=0pt\relax}
\providecommand{\BIBentryALTinterwordstretchfactor}{4}
\providecommand{\BIBentryALTinterwordspacing}{\spaceskip=\fontdimen2\font plus
\BIBentryALTinterwordstretchfactor\fontdimen3\font minus
  \fontdimen4\font\relax}
\providecommand{\BIBforeignlanguage}[2]{{%
\expandafter\ifx\csname l@#1\endcsname\relax
\typeout{** WARNING: IEEEtran.bst: No hyphenation pattern has been}%
\typeout{** loaded for the language `#1'. Using the pattern for}%
\typeout{** the default language instead.}%
\else
\language=\csname l@#1\endcsname
\fi
#2}}
\providecommand{\BIBdecl}{\relax}
\BIBdecl

\bibitem{probabilistic_robotics}
S.~Thrun, W.~Burgard, and D.~Fox, \emph{Probabilistic robotics}, ser.
  Intelligent robotics and autonomous agents.\hskip 1em plus 0.5em minus
  0.4em\relax {MIT} Press, 2005.

\bibitem{idtechex}
IDTechEx, ``Lidar 2020–2030: Technologies, players, markets and forecasts,''
  report, 2020.

\bibitem{DARPA1}
T.~Roucek, M.~Pecka, P.~C{\'{\i}}zek, T.~Petr{\'{\i}}cek, J.~Bayer,
  V.~Salansk{\'{y}}, D.~Hert, M.~Petrl{\'{\i}}k, T.~B{\'{a}}ca,
  V.~Spurn{\'{y}}, F.~Pomerleau, V.~Kubelka, J.~Faigl, K.~Zimmermann, M.~Saska,
  T.~Svoboda, and T.~Krajn{\'{\i}}k, ``{DARPA} subterranean challenge:
  Multi-robotic exploration of underground environments,'' in \emph{Proceedings
  of MESAS}, ser. Lecture Notes in Computer Science, J.~Mazal, A.~Fagiolini,
  and P.~Vas{\'{\i}}k, Eds., vol. 11995, 2019, pp. 274--290.

\bibitem{zhimon20jist}
Z.~Yan, S.~Schreiberhuber, G.~Halmetschlager, T.~Duckett, M.~Vincze, and
  N.~Bellotto, ``Robot perception of static and dynamic objects with an
  autonomous floor scrubber,'' \emph{Intelligent Service Robotics}, vol.~13,
  no.~3, pp. 403--417, 2020.

\bibitem{ls21iros}
L.~Sun, M.~Taher, C.~Wild, C.~Zhao, Y.~Zhang, F.~Majer, Z.~Yan, T.~Krajnik,
  T.~J. Prescott, and T.~Duckett, ``Robust and long-term monocular
  teach-and-repeat navigation using a single-experience map,'' in
  \emph{Proceedings of IROS}, 2021.

\bibitem{yz20iros}
Z.~Yan, L.~Sun, T.~Krajnik, and Y.~Ruichek, ``{EU} long-term dataset with
  multiple sensors for autonomous driving,'' in \emph{Proceedings of IROS},
  2020, pp. 10\,697--10\,704.

\bibitem{Rasshofer2011}
R.~H. Rasshofer, M.~Spies, and H.~Spies, ``Influences of weather phenomena on
  automotive laser radar systems,'' in \emph{Advances in {Radio} {Science}},
  vol.~9, no. B.2, 2011, pp. 49--60.

\bibitem{Ogawa2016}
T.~Ogawa and G.~Wanielik, ``{TOF}-{LIDAR} signal processing using the {CFAR}
  detector,'' in \emph{Advances in {Radio} {Science}}, vol.~14, no.~F., 2016,
  pp. 161--167.

\bibitem{you20spm}
Y.~Li and J.~Iba{\~{n}}ez{-}Guzm{\'{a}}n, ``Lidar for autonomous driving: The
  principles, challenges, and trends for automotive lidar and perception
  systems,'' \emph{IEEE Signal Processing Magazine}, vol.~37, no.~4, pp.
  50--61, 2020.

\bibitem{Lekner1988}
J.~Lekner and M.~C. Dorf, ``Why some things are darker when wet,''
  \emph{Applied Optics}, vol.~27, no.~7, pp. 1278--1280, 1988.

\bibitem{sac2022george}
G.~Broughton, J.~Janota, J.~Blaha, Z.~Yan, and T.~Krajnik, ``Bootstrapped
  learning for car detection in planar lidars,'' in \emph{Proceedings of SAC},
  2022.

\bibitem{pcl}
R.~B. Rusu and S.~Cousins, ``{3D} is here: {Point Cloud Library (PCL)},'' in
  \emph{Proceedings of ICRA}, 2011.

\bibitem{ros}
M.~Quigley, K.~Conley, B.~P. Gerkey, J.~Faust, T.~Foote, J.~Leibs, R.~Wheeler,
  and A.~Y. Ng, ``{ROS}: an open-source robot operating system,'' in \emph{ICRA
  Workshop on Open Source Software}, 2009.

\bibitem{yz17iros}
Z.~Yan, T.~Duckett, and N.~Bellotto, ``Online learning for human classification
  in {3D LiDAR-based} tracking,'' in \emph{Proceedings of IROS}, 2017, pp.
  864--871.

\bibitem{yz19auro}
------, ``Online learning for 3d lidar-based human detection: experimental
  analysis of point cloud clustering and classification methods,''
  \emph{Autonomous Robots}, vol.~44, no.~2, pp. 147--164, 2020.

\bibitem{KITTI}
A.~Geiger, P.~Lenz, C.~Stiller, and R.~Urtasun, ``Vision meets robotics: The
  {KITTI} dataset,'' \emph{International Journal of Robotics Research},
  vol.~32, no.~11, pp. 1231--1237, 2013.

\bibitem{oxford-dataset}
W.~Maddern, G.~Pascoe, C.~Linegar, and P.~Newman, ``{1 Year, 1000km: The Oxford
  RobotCar Dataset},'' \emph{The International Journal of Robotics Research},
  vol.~36, no.~1, pp. 3--15, 2017.

\bibitem{KAIST}
J.~Jeong, Y.~Cho, Y.-S. Shin, H.~Roh, and A.~Kim, ``Complex urban dataset with
  multi-level sensors from highly diverse urban environments,''
  \emph{International Journal of Robotics Research}, 2019.

\bibitem{cadcd}
M.~Pitropov, D.~E. Garcia, J.~Rebello, M.~Smart, C.~Wang, K.~Czarnecki, and
  S.~L. Waslander, ``Canadian adverse driving conditions dataset,''
  \emph{International Journal of Robotics Research}, vol.~40, no. 4-5, 2021.

\bibitem{nuScenes}
H.~Caesar, V.~Bankiti, A.~H. Lang, S.~Vora, V.~E. Liong, Q.~Xu, A.~Krishnan,
  Y.~Pan, G.~Baldan, and O.~Beijbom, ``nuscenes: {A} multimodal dataset for
  autonomous driving,'' in \emph{Proceedings of CVPR}, 2020, pp.
  11\,618--11\,628.

\bibitem{yang21itsc}
R.~Yang, Z.~Yan, T.~Yang, and Y.~Ruichek, ``Efficient online transfer learning
  for 3d object classification in autonomous driving,'' in \emph{Proceedings of
  ITSC}, 2021.

\bibitem{kidono11iv}
K.~Kidono, T.~Miyasaka, A.~Watanabe, T.~Naito, and J.~Miura, ``Pedestrian
  recognition using high-definition {LIDAR},'' in \emph{Proceedings of IV},
  2011, pp. 405--410.

\bibitem{dewan16icra}
A.~Dewan, T.~Caselitz, G.~D. Tipaldi, and W.~Burgard, ``Motion-based detection
  and tracking in {3D LiDAR} scans,'' in \emph{Proceedings of ICRA}, 2016, pp.
  4508--4513.

\bibitem{VoxelNet}
Y.~Zhou and O.~Tuzel, ``Voxelnet: End-to-end learning for point cloud based 3d
  object detection,'' in \emph{Proceedings of CVPR}, 2018, pp. 4490--4499.

\bibitem{YOLO3D}
W.~Ali, S.~Abdelkarim, M.~Zidan, M.~Zahran, and A.~E. Sallab, ``{YOLO3D:}
  end-to-end real-time 3d oriented object bounding box detection from lidar
  point cloud,'' in \emph{ECCV 2018 Workshops}, ser. Lecture Notes in Computer
  Science, L.~Leal{-}Taix{\'{e}} and S.~Roth, Eds., vol. 11131, 2018, pp.
  716--728.

\bibitem{navarro-serment09fsr}
L.~E. Navarro-Serment, C.~Mertz, and M.~Hebert, ``Pedestrian detection and
  tracking using three-dimensional ladar data,'' in \emph{Proceedings of FSR},
  2009, pp. 103--112.

\bibitem{haselich14iros}
M.~H\"aselich, B.~J\"obgen, N.~Wojke, J.~Hedrich, and D.~Paulus,
  ``Confidence-based pedestrian tracking in unstructured environments using 3d
  laser distance measurements,'' in \emph{Proceedings of IROS}, 2014, pp.
  4118--4123.

\bibitem{li16its}
K.~Li, X.~Wang, Y.~Xu, and J.~Wang, ``Density enhancement-based long-range
  pedestrian detection using 3-d range data,'' \emph{IEEE Transactions on
  Intelligent Transportation Systems}, vol.~17, pp. 1368--1380, 2016.

\bibitem{wang15rss}
D.~Z. Wang and I.~Posner, ``Voting for voting in online point cloud object
  detection,'' in \emph{Proceedings of RSS}, 2015.

\bibitem{spinello11icra}
L.~Spinello, M.~Luber, and K.~O. Arras, ``Tracking people in 3d using a
  bottom-up top-down detector,'' in \emph{Proceedings of ICRA}, 2011, pp.
  1304--1310.

\bibitem{deuge13acra}
M.~D. Deuge, A.~Quadros, C.~Hung, and B.~Douillard, ``Unsupervised feature
  learning for classification of outdoor 3d scans,'' in \emph{Proceedings of
  ACRA}, 2013.

\bibitem{teichman12ijrr}
A.~Teichman and S.~Thrun, ``Tracking-based semi-supervised learning,''
  \emph{International Journal of Robotics Research}, vol.~31, no.~7, pp.
  804--818, 2012.

\bibitem{dequaire17ijrr}
J.~Dequaire, P.~Ondruska, D.~Rao, D.~Wang, and I.~Posner, ``Deep tracking in
  the wild: End-to-end tracking using recurrent neural networks,''
  \emph{International Journal of Robotics Research}, pp. 1--21, 2017.

\bibitem{sualeh19sensor}
M.~Sualeh and G.~Kim, ``Dynamic multi-lidar based multiple object detection and
  tracking,'' \emph{Sensors}, vol.~19, no.~6, p. 1474, 2019.

\bibitem{qi21cvpr}
C.~R. Qi, Y.~Zhou, M.~Najibi, P.~Sun, K.~Vo, B.~Deng, and D.~Anguelov,
  ``Offboard 3d object detection from point cloud sequences,'' in
  \emph{Proceedings of CVPR}, 2021, pp. 6134--6144.

\bibitem{yz18iros}
Z.~Yan, L.~Sun, T.~Duckett, and N.~Bellotto, ``Multisensor online transfer
  learning for 3d lidar-based human detection with a mobile robot,'' in
  \emph{Proceedings of IROS}, 2018, pp. 7635--7640.

\bibitem{ls20icra}
L.~Sun, D.~Adolfsson, M.~Magnusson, H.~Andreasson, I.~Posner, and T.~Duckett,
  ``Localising faster: Efficient and precise lidar-based robot localisation in
  large-scale environments,'' in \emph{Proceedings of ICRA}, 2020, pp.
  4386--4392.

\bibitem{wang2020pointloc}
W.~Wang, B.~Wang, P.~Zhao, C.~Chen, R.~Clark, B.~Yang, A.~Markham, and
  N.~Trigoni, ``Pointloc: Deep pose regressor for lidar point cloud
  localization,'' \emph{arXiv preprint arXiv:2003.02392}, 2020.

\bibitem{dube2017segmatch}
R.~Dub{\'e}, D.~Dugas, E.~Stumm, J.~Nieto, R.~Siegwart, and C.~Cadena,
  ``{SegMatch}: Segment based place recognition in 3d point clouds,'' in
  \emph{Proceedings of ICRA}, 2017, pp. 5266--5272.

\bibitem{segmap2018}
R.~Dub{\'e}, A.~Cramariuc, D.~Dugas, J.~Nieto, R.~Siegwart, and C.~Cadena,
  ``{SegMap}: 3d segment mapping using data-driven descriptors,'' in
  \emph{Proceedings of RSS}, 2018.

\bibitem{tinchev2019learning}
G.~Tinchev, A.~Penate-Sanchez, and M.~Fallon, ``Learning to see the wood for
  the trees: Deep laser localization in urban and natural environments on a
  cpu,'' \emph{IEEE Robotics and Automation Letters}, vol.~4, no.~2, pp.
  1327--1334, 2019.

\bibitem{kong2020semantic}
X.~Kong, X.~Yang, G.~Zhai, X.~Zhao, X.~Zeng, M.~Wang, Y.~Liu, W.~Li, and
  F.~Wen, ``Semantic graph based place recognition for 3d point clouds,''
  \emph{arXiv preprint arXiv:2008.11459}, 2020.

\bibitem{he2016m2dp}
L.~He, X.~Wang, and H.~Zhang, ``M2dp: A novel 3d point cloud descriptor and its
  application in loop closure detection,'' in \emph{Proceedings of IROS}, 2016,
  pp. 231--237.

\bibitem{kim2018scan}
G.~Kim and A.~Kim, ``Scan context: Egocentric spatial descriptor for place
  recognition within 3d point cloud map,'' in \emph{Proceedings of IROS}, 2018,
  pp. 4802--4809.

\bibitem{cop2018delight}
K.~P. Cop, P.~V. Borges, and R.~Dub{\'e}, ``Delight: An efficient descriptor
  for global localisation using lidar intensities,'' in \emph{Proceedings of
  ICRA}, 2018, pp. 3653--3660.

\bibitem{kim20191}
G.~Kim, B.~Park, and A.~Kim, ``1-day learning, 1-year localization: Long-term
  lidar localization using scan context image,'' \emph{IEEE Robotics and
  Automation Letters}, vol.~4, no.~2, pp. 1948--1955, 2019.

\bibitem{chen2020overlapnet}
X.~Chen, T.~L{\"a}be, A.~Milioto, T.~R{\"o}hling, O.~Vysotska, A.~Haag,
  J.~Behley, C.~Stachniss, and F.~Fraunhofer, ``Overlapnet: Loop closing for
  lidar-based slam,'' in \emph{Proceedings of RSS}, 2020.

\bibitem{angelina2018pointnetvlad}
M.~Angelina~Uy and G.~Hee~Lee, ``Pointnetvlad: Deep point cloud based retrieval
  for large-scale place recognition,'' in \emph{Proceedings of CVPR}, 2018, pp.
  4470--4479.

\bibitem{liu2019lpd}
Z.~Liu, S.~Zhou, C.~Suo, P.~Yin, W.~Chen, H.~Wang, H.~Li, and Y.-H. Liu,
  ``Lpd-net: 3d point cloud learning for large-scale place recognition and
  environment analysis,'' in \emph{Proceedings of ICCV}, 2019, pp. 2831--2840.

\bibitem{kucnerconditional}
T.~Kucner, J.~Saarinen, M.~Magnusson, and A.~J.~Lilienthal, ``Conditional
  transition maps: Learning motion patterns in dynamic environments,'' in
  \emph{Proceedings of IROS}, 2013.

\bibitem{pomerleau2014long}
F.~Pomerleau, P.~Kr{\"u}si, F.~Colas, P.~Furgale, and R.~Siegwart, ``Long-term
  3d map maintenance in dynamic environments,'' in \emph{Proceedings of ICRA},
  2014, pp. 3712--3719.

\bibitem{ls18icra}
L.~Sun, Z.~Yan, S.~M. Mellado, M.~Hanheide, and T.~Duckett, ``{3DOF} pedestrian
  trajectory prediction learned from long-term autonomous mobile robot
  deployment data,'' in \emph{Proceedings of ICRA}, 2018, pp. 1--7.

\bibitem{ls18ral}
L.~Sun, Z.~Yan, A.~Zaganidis, C.~Zhao, and T.~Duckett, ``Recurrent-octomap:
  Learning state-based map refinement for long-term semantic mapping with
  3d-lidar data,'' \emph{IEEE Robotics and Automation Letters}, vol.~3, no.~4,
  pp. 3749--3756, 2018.

\bibitem{vintr19ecmr}
T.~Vintr, T.~Krajnik, S.~Molina, R.~Senanayake, G.~Broughton, Z.~Yan,
  J.~Ulrich, T.~Kucner, C.~Swaminathan, F.~Majer, M.~Stachova, and
  A.~Lilienthal, ``Time-varying pedestrian flow models for service robots,'' in
  \emph{Proceedings of ECMR}, 2019, pp. 1--7.

\bibitem{vintr19icra}
T.~Vintr, Z.~Yan, T.~Duckett, and T.~Krajnik, ``Spatio-temporal representation
  for long-term anticipation of human presence in service robotics,'' in
  \emph{Proceedings of ICRA}, 2019, pp. 2620--2626.

\bibitem{vintr20iros}
T.~Vintr, Z.~Yan, F.~Kubis, J.~Blaha, J.~Ulrich, F.~K. Eyisoy, C.~S.
  Swaminathan, S.~Molina, T.~P. Kucner, M.~Magnusson, G.~Cielniak, J.~Faigl,
  T.~Duckett, A.~J. Lilienthal, and T.~Krajnik, ``Evaluating flow models for
  socially compliant robot navigation in populated environments,'' in
  \emph{Proceedings of IROS}, 2020, pp. 11\,197--11\,204.

\bibitem{broughton2020learning}
G.~Broughton, F.~Majer, T.~Rou{\v{c}}ek, Y.~Ruichek, Z.~Yan, and
  T.~Krajn{\'\i}k, ``Learning to see through the haze: Multi-sensor
  learning-fusion system for vulnerable traffic participant detection in fog,''
  \emph{Robotics and Autonomous Systems}, vol. 136, p. 103687, 2021.

\bibitem{roy2020physical}
G.~Roy, X.~Cao, R.~Bernier, and G.~Tremblay, ``Physical model of snow
  precipitation interaction with a 3d lidar scanner,'' \emph{Applied Optics},
  vol.~59, no.~25, pp. 7660--7669, 2020.

\bibitem{yang20iros}
T.~Yang, Y.~Li, Y.~Ruichek, and Z.~Yan, ``{LaNoising}: A data-driven approach
  for {903nm} {ToF} {LiDAR} performance modeling under fog,'' in
  \emph{Proceedings of IROS}, 2020, pp. 10\,084--10\,091.

\bibitem{hahner2021fog}
M.~Hahner, C.~Sakaridis, D.~Dai, and L.~V. Gool, ``Fog simulation on real lidar
  point clouds for 3d object detection in adverse weather,'' 2021.

\bibitem{8575761}
N.~{Charron}, S.~{Phillips}, and S.~L. {Waslander}, ``De-noising of lidar point
  clouds corrupted by snowfall,'' in \emph{Proceedings of CRV}, 2018, pp.
  254--261.

\bibitem{heinzler2020cnn}
R.~Heinzler, F.~Piewak, P.~Schindler, and W.~Stork, ``Cnn-based lidar point
  cloud de-noising in adverse weather,'' \emph{IEEE Robotics and Automation
  Letters}, vol.~5, no.~2, pp. 2514--2521, 2020.

\bibitem{park2020fast}
J.-I. Park, J.~Park, and K.-S. Kim, ``Fast and accurate desnowing algorithm for
  lidar point clouds,'' \emph{IEEE Access}, vol.~8, pp. 160\,202--160\,212,
  2020.

\bibitem{kurup2021dsor}
A.~Kurup and J.~Bos, ``Dsor: A scalable statistical filter for removing falling
  snow from lidar point clouds in severe winter weather,'' \emph{arXiv preprint
  arXiv:2109.07078}, 2021.

\bibitem{zermas17icra}
D.~Zermas, I.~Izzat, and N.~Papanikolopoulos, ``Fast segmentation of 3d point
  clouds: A paradigm on lidar data for autonomous vehicle applications,'' in
  \emph{Proceedings of ICRA}, 2017.

\bibitem{bogoslavskyi16iros}
I.~Bogoslavskyi and C.~Stachniss, ``Fast range image-based segmentation of
  sparse 3d laser scans for online operation,'' in \emph{Proceedings of IROS},
  2016, pp. 163--169.

\bibitem{autoware}
S.~Kato, S.~Tokunaga, Y.~Maruyama, S.~Maeda, M.~Hirabayashi, Y.~Kitsukawa,
  A.~Monrroy, T.~Ando, Y.~Fujii, and T.~Azumi, ``Autoware on board: enabling
  autonomous vehicles with embedded systems,'' in \emph{Proceedings of ICCPS},
  2018, pp. 287--296.

\bibitem{insclustering}
Y.~Li, C.~L. Bihan, T.~Pourtau, and T.~Ristorcelli, ``{InsClustering}:
  Instantly clustering lidar range measures for autonomous vehicle,'' in
  \emph{Proceedings of ITSC}, 2020, pp. 1--6.

\bibitem{shackleton10avss}
J.~Shackleton, B.~V. Voorst, and J.~A. Hesch, ``Tracking people with a
  360-degree lidar,'' in \emph{Proceedings of the Seventh IEEE International
  Conference on Advanced Video and Signal Based Surveillance (AVSS)}, 2010, pp.
  420--426.

\bibitem{RusuDissertation}
R.~B. Rusu, ``Semantic {3D} object maps for everyday manipulation in human
  living environments,'' Ph.D. dissertation, Computer Science department,
  Technische Universitaet Muenchen, Germany, 2009.

\bibitem{himmelsbach2010fast}
M.~Himmelsbach, F.~V. Hundelshausen, and H.-J. Wuensche, ``Fast segmentation of
  3d point clouds for ground vehicles,'' in \emph{2010 IEEE Intelligent
  Vehicles Symposium}, 2010, pp. 560--565.

\bibitem{Complex-yolo-2018}
M.~Simon, S.~Milz, K.~Amende, and H.~Gross, ``Complex-yolo: An
  euler-region-proposal for real-time 3d object detection on point clouds,'' in
  \emph{ECCV Workshops}, L.~Leal{-}Taix{\'{e}} and S.~Roth, Eds., 2018.

\bibitem{Pixor-2018}
B.~Yang, W.~Luo, and R.~Urtasun, ``{PIXOR:} real-time 3d object detection from
  point clouds,'' in \emph{Proceedings of CVPR}, 2018, pp. 7652--7660.

\bibitem{Pointpillars-2019}
A.~H. Lang, S.~Vora, H.~Caesar, L.~Zhou, J.~Yang, and O.~Beijbom,
  ``Pointpillars: Fast encoders for object detection from point clouds,'' in
  \emph{Proceedings of CVPR}, 2019, pp. 12\,697--12\,705.

\bibitem{qi2017pointnet}
C.~R. Qi, H.~Su, K.~Mo, and L.~J. Guibas, ``Pointnet: Deep learning on point
  sets for 3d classification and segmentation,'' in \emph{Proceedings of CVPR},
  2017, pp. 652--660.

\bibitem{ls21icra}
Z.~Zhou, C.~Zhao, D.~Adolfsson, S.~Su, Y.~Gao, T.~Duckett, and L.~Sun,
  ``Ndt-transformer: Large-scale 3d point cloud localisation using the normal
  distribution transform representation,'' in \emph{Proceedings of ICRA}, 2021.

\bibitem{ryde20103d}
J.~Ryde and H.~Hu, ``3d mapping with multi-resolution occupied voxel lists,''
  \emph{Autonomous Robots}, vol.~28, no.~2, p. 169, 2010.

\bibitem{thrun_monte_2000}
S.~Thrun, D.~Fox, W.~Burgard, and {others}, ``{Monte Carlo }localization with
  mixture proposal distribution,'' in \emph{{AAAI}/{IAAI}}, 2000, pp. 859--865.

\bibitem{he2013observation}
T.~He and S.~Hirose, ``Observation-driven bayesian filtering for global
  location estimation in the field area,'' \emph{Journal of Field Robotics},
  vol.~30, no.~4, pp. 489--518, 2013.

\bibitem{kucner-2015-mcl}
T.~P. Kucner, M.~Magnusson, and A.~J. Lilienthal, ``Where am {I}?: An
  {NDT}-based prior for {MCL},'' in \emph{Proceedings of ECMR}, 2015.

\bibitem{seow2017detecting}
Y.~Seow, R.~Miyagusuku, A.~Yamashita, and H.~Asama, ``Detecting and solving the
  kidnapped robot problem using laser range finder and wifi signal,'' in
  \emph{Proceedings of RCAR}, 2017, pp. 303--308.

\bibitem{oh_map-based_2004}
S.~M. Oh, S.~Tariq, B.~N. Walker, and F.~Dellaert, ``Map-based priors for
  localization,'' in \emph{Proceedings of IROS}, vol.~3, 2004, pp. 2179--2184.

\bibitem{akai2020hybrid}
N.~Akai, T.~Hirayama, and H.~Murase, ``Hybrid localization using model-and
  learning-based methods: Fusion of monte carlo and e2e localizations via
  importance sampling,'' in \emph{Proceedings of ICRA}, 2020.

\bibitem{Arandjelovic_2016_CVPR}
R.~Arandjelovic, P.~Gronat, A.~Torii, T.~Pajdla, and J.~Sivic, ``Netvlad: Cnn
  architecture for weakly supervised place recognition,'' in \emph{Proceedings
  of CVPR}, June 2016.

\bibitem{wang2019dynamic}
Y.~Wang, Y.~Sun, Z.~Liu, S.~E. Sarma, M.~M. Bronstein, and J.~M. Solomon,
  ``Dynamic graph cnn for learning on point clouds,'' \emph{{ACM} Transactions
  On Graphics (tog)}, vol.~38, no.~5, pp. 1--12, 2019.

\bibitem{agricultural}
T.~Duckett, S.~Pearson, S.~Blackmore, and B.~Grieve, ``Agricultural robotics:
  The future of robotic agriculture,'' \emph{CoRR}, vol. abs/1806.06762, 2018.

\bibitem{DARPA2}
T.~Rou{\v{c}}ek, M.~Pecka, P.~{\v{C}}{\'\i}{\v{z}}ek,
  T.~Pet{\v{r}}{\'\i}{\v{c}}ek, J.~Bayer, V.~{\v{S}}alansk{\`y}, T.~Azayev,
  D.~He{\v{r}}t, M.~Petrl{\'\i}k, T.~B{\'a}{\v{c}}a \emph{et~al.}, ``System for
  multi-robotic exploration of underground environments ctu-cras-norlab in the
  darpa subterranean challenge,'' \emph{arXiv preprint arXiv:2110.05911}, 2021.

\bibitem{biber}
P.~Biber and T.~Duckett, ``Dynamic maps for long-term operation of mobile
  service robots,'' in \emph{Proceedings of RSS}, 2005, pp. 17--24.

\bibitem{tipaldi}
G.~D. Tipaldi, D.~Meyer-Delius, and W.~Burgard, ``Lifelong localization in
  changing environments,'' \emph{International Journal of Robotics Research},
  vol.~32, no.~14, pp. 1662--1678, 2013.

\bibitem{krajnik2016persistent}
T.~Krajn{\'\i}k, J.~P. Fentanes, M.~Hanheide, and T.~Duckett, ``Persistent
  localization and life-long mapping in changing environments using the
  frequency map enhancement,'' in \emph{Proceedings of IROS}, 2016, pp.
  4558--4563.

\bibitem{cliff}
C.~S. Swaminathan, T.~P. Kucner, M.~Magnusson, L.~Palmieri, and A.~J.
  Lilienthal, ``Down the cliff: Flow-aware trajectory planning under motion
  pattern uncertainty,'' in \emph{Proceedings of IROS}, 2018, pp. 7403--7409.

\bibitem{ayoung}
G.~Kim and A.~Kim, ``{LT-mapper}: A modular framework for lidar-based lifelong
  mapping,'' in \emph{Proceedings of ICRA}, 2022.

\bibitem{fremen}
T.~Krajn{\'\i}k, J.~P. Fentanes, J.~M. Santos, and T.~Duckett, ``Fremen:
  Frequency map enhancement for long-term mobile robot autonomy in changing
  environments,'' \emph{IEEE Transactions on Robotics}, vol.~33, no.~4, pp.
  964--977, 2017.

\bibitem{tomek}
T.~P. Kucner, A.~J. Lilienthal, M.~Magnusson, L.~Palmieri, and C.~S.
  Swaminathan, \emph{Probabilistic mapping of spatial motion patterns for
  mobile robots}.\hskip 1em plus 0.5em minus 0.4em\relax Springer, 2020.

\bibitem{Peynot2009}
T.~{Peynot}, J.~{Underwood}, and S.~{Scheding}, ``Towards reliable perception
  for unmanned ground vehicles in challenging conditions,'' in
  \emph{Proceedings of IROS}, 2009, pp. 1170--1176.

\bibitem{michaud2015towards}
S.~Michaud, J.-F. Lalonde, and P.~Giguere, ``Towards characterizing the
  behavior of lidars in snowy conditions,'' in \emph{IROS PPNIV Workshop},
  vol.~28, 2015.

\bibitem{Hasirlioglu2017}
S.~Hasirlioglu, A.~Riener, W.~Huber, and P.~Wintersberger, ``Effects of exhaust
  gases on laser scanner data quality at low ambient temperatures,'' in
  \emph{Proceedings of IV)}, 2017, pp. 1708--1713.

\bibitem{Ashraf2018}
I.~Ashraf and Y.~Park, ``Effects of fog attenuation on lidar data in urban
  environment,'' in \emph{Smart Photonic and Optoelectronic Integrated Circuits
  XX}, vol. 10536, 2018, p. 1053623.

\bibitem{Filgueira2017}
A.~Filgueira, H.~Gonz{\'a}lez-Jorge, S.~Lag{\"u}ela, L.~D{\'\i}az-Vilari{\~n}o,
  and P.~Arias, ``Quantifying the influence of rain in lidar performance,''
  \emph{Measurement}, vol.~95, pp. 143--148, 2017.

\bibitem{goodin2019predicting}
C.~Goodin, D.~Carruth, M.~Doude, and C.~Hudson, ``Predicting the influence of
  rain on lidar in adas,'' \emph{Electronics}, vol.~8, no.~1, p.~89, 2019.

\bibitem{yoneda2019automated}
K.~Yoneda, N.~Suganuma, R.~Yanase, and M.~Aldibaja, ``Automated driving
  recognition technologies for adverse weather conditions,'' \emph{IATSS
  research}, vol.~43, no.~4, pp. 253--262, 2019.

\bibitem{Kutila2018}
M.~Kutila, P.~Pyyk{\"o}nen, H.~Holzh{\"u}ter, M.~Colomb, and P.~Duthon,
  ``Automotive lidar performance verification in fog and rain,'' in
  \emph{Proceedings of ITSC}, 2018, pp. 1695--1701.

\bibitem{jokela2019testing}
M.~Jokela, M.~Kutila, and P.~Pyyk{\"o}nen, ``Testing and validation of
  automotive point-cloud sensors in adverse weather conditions,'' \emph{Applied
  Sciences}, vol.~9, no.~11, p. 2341, 2019.

\bibitem{Karl2021}
K.~Montalban, C.~Reymann, D.~Atchuthan, P.-E. Dupouy, N.~Riviere, and
  S.~Lacroix, ``A quantitative analysis of point clouds from automotive lidars
  exposed to artificial rain and fog,'' \emph{Atmosphere}, vol.~12, no.~6,
  2021.

\bibitem{Alexander2020}
A.~Carballo, J.~Lambert, A.~Monrroy, D.~Wong, P.~Narksri, Y.~Kitsukawa,
  E.~Takeuchi, S.~Kato, and K.~Takeda, ``Libre: The multiple 3d lidar
  dataset,'' in \emph{Proceedings of IV}, 2020, pp. 1094--1101.

\bibitem{you20fog}
Y.~{Li}, P.~{Duthon}, M.~{Colomb}, and J.~{Ibanez-Guzman}, ``What happens for a
  tof lidar in fog?'' \emph{IEEE Transactions on Intelligent Transportation
  Systems}, pp. 1--12, 2020.

\bibitem{Bijelic2018}
M.~Bijelic, T.~Gruber, and W.~Ritter, ``A {Benchmark} for {Lidar} {Sensors} in
  {Fog}: {Is} {Detection} {Breaking} {Down}?'' in \emph{Proceedings of IV},
  2018, pp. 760--767.

\bibitem{yang21tits}
T.~Yang, Y.~Li, Y.~Ruichek, and Z.~Yan, ``Performance modeling a near-infrared
  {ToF LiDAR} under fog: A data-driven approach,'' \emph{IEEE Transactions on
  Intelligent Transportation Systems}, 2021.

\bibitem{Yin2006}
S.~Yin and W.~Wang, ``Lidar signal denoising based on wavelet domain spatial
  filtering,'' in \emph{Proceedings of RADAR}, 2006, pp. 1--3.

\bibitem{Sun2017}
L.~Sun, Z.~Dong, R.~Zhang, R.~Fan, and D.~Chen, ``Waveform lidar signal
  denoising based on connected domains,'' \emph{Frontiers of Optoelectronics},
  vol.~10, no.~4, pp. 388--394, 2017.

\bibitem{Azadbakht2013}
M.~Azadbakht, C.~S. Fraser, C.~Zhang, and J.~Leach, ``A signal denoising method
  for full-waveform lidar data,'' \emph{Proceedings of ISPRS Annals}, pp.
  11--13, 2013.

\bibitem{Nie2011}
J.~Nie, Y.~Hu, and A.~Zi, ``Outlier detection of scattered point cloud by
  classification,'' \emph{Journal of Computer-Aided Design \& Computer
  Graphics}, vol.~9, no.~23, pp. 1526--1532, 2011.

\bibitem{Cao2013}
S.~Cao, J.~Yue, and W.~Ma, ``Bilateral filtering denoise algorithm for point
  cloud based on feature selection,'' \emph{Journal of Southeast University
  (Natural Science Edition)}, vol.~43, no.~S2, pp. 351--354, 2013.

\bibitem{Zhu2015}
J.~Zhu, X.~Hu, Z.~Zhang, and X.~Xiong, ``Hierarchical outlier detection for
  point cloud data using a density analysis method [j],'' \emph{Acta Geodaetica
  et Cartographica Sinica}, vol.~44, no.~3, pp. 282--290, 2015.

\bibitem{Javaheri2017}
A.~Javaheri, C.~Brites, F.~Pereira, and J.~Ascenso, ``Subjective and objective
  quality evaluation of 3d point cloud denoising algorithms,'' in
  \emph{Proceedings of ICMEW}, 2017, pp. 1--6.

\bibitem{Yonghua2017DenoisingAO}
S.~Yong-hua, Z.~Xuqing, N.~Xuefeng, Y.~Guodong, and Z.~Ji-kai, ``Denoising
  algorithm of airborne lidar point cloud based on 3d grid,''
  \emph{International Journal of Signal Processing, Image Processing and
  Pattern Recognition}, vol.~10, pp. 85--92, 2017.

\bibitem{Holz2015}
D.~Holz, A.~E. Ichim, F.~Tombari, R.~B. Rusu, and S.~Behnke, ``Registration
  with the point cloud library: A modular framework for aligning in 3-d,''
  \emph{IEEE Robotics \& Automation Magazine}, vol.~22, no.~4, pp. 110--124,
  2015.

\bibitem{duan2021low}
Y.~Duan, C.~Yang, H.~Chen, W.~Yan, and H.~Li, ``Low-complexity point cloud
  denoising for lidar by pca-based dimension reduction,'' \emph{Optics
  Communications}, vol. 482, p. 126567, 2021.

\bibitem{Waymo}
P.~Sun, H.~Kretzschmar, X.~Dotiwalla, A.~Chouard, V.~Patnaik, P.~Tsui, J.~Guo,
  Y.~Zhou, Y.~Chai, B.~Caine, V.~Vasudevan, W.~Han, J.~Ngiam, H.~Zhao,
  A.~Timofeev, S.~Ettinger, M.~Krivokon, A.~Gao, A.~Joshi, Y.~Zhang, J.~Shlens,
  Z.~Chen, and D.~Anguelov, ``Scalability in perception for autonomous driving:
  Waymo open dataset,'' in \emph{Proceedings of CVPR}, 2020, pp. 2446--2454.

\bibitem{gruber2019pixel}
T.~Gruber, M.~Bijelic, F.~Heide, W.~Ritter, and K.~Dietmayer, ``Pixel-accurate
  depth evaluation in realistic driving scenarios,'' in \emph{Proceedings of
  3DV}, 2019, pp. 95--105.

\bibitem{Gruber_2019_ICCV}
T.~Gruber, F.~Julca-Aguilar, M.~Bijelic, and F.~Heide, ``Gated2depth: Real-time
  dense lidar from gated images,'' in \emph{Proceedings of ICCV}, 2019.

\bibitem{Bijelic20}
M.~{Bijelic}, T.~{Gruber}, F.~{Mannan}, F.~{Kraus}, W.~{Ritter},
  K.~{Dietmayer}, and F.~{Heide}, ``Seeing through fog without seeing fog: Deep
  multimodal sensor fusion in unseen adverse weather,'' in \emph{Proceedings of
  CVPR}, 2020, pp. 11\,679--11\,689.

\bibitem{zhang21iros}
C.~Zhang, Z.~Huang, M.~H. Ang~Jr, and D.~Rus, ``Lidar degradation
  quantification for autonomous driving in rain,'' in \emph{Proceedings of
  IROS}, 2021.

\bibitem{sheeny2020radiate}
M.~Sheeny, E.~De~Pellegrin, S.~Mukherjee, A.~Ahrabian, S.~Wang, and A.~Wallace,
  ``Radiate: A radar dataset for automotive perception,'' \emph{arXiv preprint
  arXiv:2010.09076}, vol.~3, no.~4, p.~7, 2020.

\bibitem{premebida14iros}
C.~Premebida, J.~Carreira, J.~Batista, and U.~Nunes, ``Pedestrian detection
  combining {RGB} and dense {LIDAR} data,'' in \emph{Proceedings of IROS},
  2014, pp. 4112--4117.

\bibitem{gonzalez15iv}
A.~Gonz{\'a}lez, G.~Villalonga, J.~Xu, D.~V{\'a}zquez, J.~Amores, and A.~M.
  L{\'o}pez, ``Multiview random forest of local experts combining {RGB} and
  {LIDAR} data for pedestrian detection,'' in \emph{Proceedings of IV}, 2015,
  pp. 356--361.

\bibitem{iliad}
T.~Linder, N.~Vaskevicius, R.~Schirmer, and K.~O. Arras, ``Cross-modal analysis
  of human detection for robotics: An industrial case study,'' in
  \emph{Proceedings of IROS}, 2021.

\bibitem{majer19ecmr}
F.~Majer, Z.~Yan, G.~Broughton, Y.~Ruichek, and T.~Krajnik, ``Learning to see
  through haze: Radar-based human detection for adverse weather conditions,''
  in \emph{Proceedings of ECMR}, 2019, pp. 1--7.

\bibitem{montemerlo08jfr}
M.~Montemerlo, J.~Becker, S.~Bhat, H.~Dahlkamp, D.~Dolgov, S.~Ettinger,
  D.~H{\"{a}}hnel, T.~Hilden, G.~Hoffmann, B.~Huhnke, D.~Johnston, S.~Klumpp,
  D.~Langer, A.~Levandowski, J.~Levinson, J.~Marcil, D.~Orenstein, J.~Paefgen,
  I.~Penny, A.~Petrovskaya, M.~Pflueger, G.~Stanek, D.~Stavens, A.~Vogt, and
  S.~Thrun, ``Junior: The stanford entry in the urban challenge,''
  \emph{Journal of Field Robotics}, vol.~25, no.~9, pp. 569--597, 2008.

\bibitem{hsr}
T.~Yamamoto, K.~Terada, A.~Ochiai, F.~Saito, Y.~Asahara, and K.~Murase,
  ``Development of human support robot as the research platform of a domestic
  mobile manipulator,'' \emph{ROBOMECH Journal}, vol.~6, no.~4, 2019.

\bibitem{STRANDS}
N.~Hawes, C.~Burbridge, F.~Jovan, L.~Kunze, B.~Lacerda, L.~Mudrov{\'{a}},
  J.~Young, J.~L. Wyatt, D.~Hebesberger, T.~K{\"{o}}rtner, R.~Ambrus, N.~Bore,
  J.~Folkesson, P.~Jensfelt, L.~Beyer, A.~Hermans, B.~Leibe, A.~Aldoma,
  T.~Faulhammer, M.~Zillich, M.~Vincze, E.~Chinellato, M.~Al{-}Omari,
  P.~Duckworth, Y.~Gatsoulis, D.~C. Hogg, A.~G. Cohn, C.~Dondrup, J.~P.
  Fentanes, T.~Krajn{\'{\i}}k, J.~M. Santos, T.~Duckett, and M.~Hanheide, ``The
  {STRANDS} project: Long-term autonomy in everyday environments,''
  \emph{{IEEE} Robotics Automation Magazine}, vol.~24, no.~3, pp. 146--156,
  2017.

\bibitem{ENRICHME}
S.~Cosar, M.~Fern{\'{a}}ndez{-}Carmona, R.~Agrigoroaie, J.~Pag{\`{e}}s,
  F.~Ferland, F.~Zhao, S.~Yue, N.~Bellotto, and A.~Tapus, ``{ENRICHME:}
  perception and interaction of an assistive robot for the elderly at home,''
  \emph{International Journal of Social Robotics}, vol.~12, no.~3, pp.
  779--805, 2020.

\bibitem{safety}
{Aptiv, AUDI, Baidu, BMW, Continental, Daimler, FCA, HERE, Infineon, Intel, and
  Volkswagen}, ``Safety first for automated driving,'' Withe Paper, Tech. Rep.,
  2019.

\bibitem{sathyamoorthy2020covid}
A.~J. Sathyamoorthy, U.~Patel, Y.~A. Savle, M.~Paul, and D.~Manocha,
  ``Covid-robot: Monitoring social distancing constraints in crowded
  scenarios,'' \emph{arXiv preprint arXiv:2008.06585}, 2020.

\bibitem{loey2021hybrid}
M.~Loey, G.~Manogaran, M.~H.~N. Taha, and N.~E.~M. Khalifa, ``A hybrid deep
  transfer learning model with machine learning methods for face mask detection
  in the era of the covid-19 pandemic,'' \emph{Measurement}, vol. 167, p.
  108288, 2021.

\bibitem{mcginn2021exploring}
C.~McGinn, R.~Scott, N.~Donnelly, K.~L. Roberts, M.~Bogue, C.~Kiernan, and
  M.~Beckett, ``Exploring the applicability of robot-assisted uv disinfection
  in radiology,'' \emph{Frontiers in Robotics and AI}, p. 193, 2021.

\bibitem{chen2020containing}
B.~Chen, S.~Marvin, and A.~While, ``Containing covid-19 in china: Ai and the
  robotic restructuring of future cities,'' \emph{Dialogues in Human
  Geography}, vol.~10, no.~2, pp. 238--241, 2020.

\bibitem{yang2020combating}
G.-Z. Yang, B.~J.~Nelson, R.~R. Murphy, H.~Choset, H.~Christensen,
  S.~H.~Collins, P.~Dario, K.~Goldberg, K.~Ikuta, N.~Jacobstein \emph{et~al.},
  ``Combating covid-19—the role of robotics in managing public health and
  infectious diseases,'' \emph{Science Robotics}, vol.~5, no.~40, p. eabb5589,
  2020.

\end{thebibliography}

\end{document}